\documentclass[10pt,journal,compsoc]{IEEEtran}
\usepackage{booktabs} 

\usepackage[ruled]{algorithm2e} 

\usepackage{overpic}
\usepackage{bbold} 
\usepackage{amsfonts} 
\usepackage{amsmath} 
\usepackage{pgfplots}
\usepackage{pgf,tikz}
\usepackage{tkz-euclide}
\usepackage{bm}
\usepackage[verbose]{wrapfig}
\usepackage{multirow}
\usepackage{mathtools}

\usepackage[utf8]{inputenc}
\usepackage{pgfplots}
\DeclareUnicodeCharacter{2212}{−}
\usepgfplotslibrary{groupplots,dateplot}
\usetikzlibrary{patterns,shapes.arrows}
\pgfplotsset{compat=newest}

\usepackage{verbatim}
\usepackage[mode=buildnew]{standalone}
\usepackage[normalem]{ulem}

\renewcommand{\phi}{\varphi}

\newcommand{\thename}{\textsc{MarioNet}}
\newcommand{\systemname}{\textsc{Neural Marionette}}
\newcommand{\dtname}{\textit{\textsc{BABEL-MAG}}}

\newcommand{\mybrac}[1]{
\big\{\,#1\,\big\}}

\usepackage{soul}
\usepackage{listings}
\usepackage{color, colortbl} 

\usepackage{enumitem}

\DeclareFontFamily{U}{mathx}{\hyphenchar\font45}
\DeclareFontShape{U}{mathx}{m}{n}{<-> mathx10}{}
\DeclareSymbolFont{mathx}{U}{mathx}{m}{n}
\DeclareMathAccent{\widebar}{0}{mathx}{"73}

\input{insbox.tex}

\usepackage{tipa}
\UndeclareTextCommand{\!}{T3}
\DeclareTextCommand{\tipaEXCLAM}{T3}{}
\DeclareRobustCommand{\!}{%
  \ifmmode\mskip-\thinmuskip\else\expandafter\tipaEXCLAM\fi
}

\usepackage{booktabs}       
\usepackage{mathrsfs}
\usetikzlibrary{arrows}
\usetikzlibrary{matrix}
\usetikzlibrary{positioning,calc,fadings}
\usetikzlibrary{backgrounds, fit}
\pgfplotsset{compat=1.14}
\usepackage{pgfplotstable}

\newcommand*{\Scale}[2][4]{\scalebox{#1}{$#2$}}%

\usepackage{pifont}

\definecolor{mygreen}{RGB}{28,172,0} 
\definecolor{mylilas}{RGB}{170,55,241}
\definecolor{ffzzcc}{rgb}{1,0.6,0.8}
\definecolor{myblue}{RGB}{23, 195, 178}
\definecolor{myyellow}{RGB}{255, 186, 8}
\definecolor{mygray}{rgb}{0.5,0.5,0.5}
\definecolor{myred}{RGB}{254, 109, 115}

\definecolor{ao}{rgb}{0.0, 0.0, 1.0}
\definecolor{airforceblue}{rgb}{0.36, 0.54, 0.66}
\definecolor{ceruleanblue}{rgb}{0.16, 0.32, 0.75}
\definecolor{cerulean}{rgb}{0.0, 0.48, 0.65}
\definecolor{celestialblue}{rgb}{0.29, 0.59, 0.82}
\definecolor{azure(colorwheel)}{rgb}{0.0, 0.5, 1.0}
\definecolor{babyblue}{rgb}{0.54, 0.81, 0.94}
\definecolor{babyblueeyes}{rgb}{0.63, 0.79, 0.95}
\definecolor{ballblue}{rgb}{0.13, 0.67, 0.8}

\definecolor{asparagus}{rgb}{0.53, 0.66, 0.42}
\definecolor{ao(english)}{rgb}{0.0, 0.5, 0.0}
\definecolor{applegreen}{rgb}{0.55, 0.71, 0.0}
\definecolor{armygreen}{rgb}{0.29, 0.33, 0.13}

\definecolor{amethyst}{rgb}{0.6, 0.4, 0.8}
\definecolor{antiquefuchsia}{rgb}{0.57, 0.36, 0.51}
\definecolor{blue-violet}{rgb}{0.54, 0.17, 0.89}
\definecolor{brightlavender}{rgb}{0.75, 0.58, 0.89}
\definecolor{brightube}{rgb}{0.82, 0.62, 0.91}
\definecolor{brilliantlavender}{rgb}{0.96, 0.73, 1.0}

\definecolor{amber}{rgb}{1.0, 0.75, 0.0}
\definecolor{amber(sae/ece)}{rgb}{1.0, 0.49, 0.0}
\definecolor{atomictangerine}{rgb}{1.0, 0.6, 0.4}
\definecolor{burntorange}{rgb}{0.8, 0.33, 0.0}
\definecolor{burntsienna}{rgb}{0.91, 0.45, 0.32}
\definecolor{cadmiumorange}{rgb}{0.93, 0.53, 0.18}
\definecolor{carrotorange}{rgb}{0.93, 0.57, 0.13}
\definecolor{chocolate(web)}{rgb}{0.82, 0.41, 0.12}
\definecolor{chromeyellow}{rgb}{1.0, 0.65, 0.0}
\definecolor{darkorange}{rgb}{1.0, 0.55, 0.0}
\definecolor{darktangerine}{rgb}{1.0, 0.66, 0.07}
\definecolor{deepcarrotorange}{rgb}{0.91, 0.41, 0.17}
\definecolor{deepsaffron}{rgb}{1.0, 0.6, 0.2}
\definecolor{fulvous}{rgb}{0.86, 0.52, 0.0}

\newcommand{\TODO}[1]{\textcolor{red}{}}
\newcommand{\INSERT}[1]{\textcolor{red}{[INSERT]}}
\newcommand{\CITE}[1]{\textcolor{red}{[CITE]}}
\usepackage[normalem]{ulem}

\newcommand\wqst{\bgroup\markoverwith{\textcolor{asparagus}{\rule[0.5ex]{2pt}{0.4pt}}}\ULon}

\newcommand\xfst{\bgroup\markoverwith{\textcolor{purple}{\rule[0.5ex]{2pt}{0.4pt}}}\ULon}

\newcommand{\dknote}[1]{{\small{\bfseries{\textcolor{cerulean}{}}}}}

\graphicspath{{figures/}}

\ifCLASSOPTIONcompsoc
  \usepackage[nocompress]{cite}
\else
  \usepackage{cite}
\fi

\ifCLASSINFOpdf
\else
\fi
\ifCLASSOPTIONcompsoc
 \usepackage[caption=false,font=footnotesize,labelfont=sf,textfont=sf]{subfig}
\else
 \usepackage[caption=false,font=footnotesize]{subfig}
\fi
\usepackage{url}

\begin{document}
%
\title{\systemname: A Transformer-based Multi-action Human Motion Synthesis System}
\author{Weiqiang~Wang \IEEEauthorrefmark{2},
         Xuefei~Zhe \IEEEauthorrefmark{2},  
         Qiuhong~Ke, 
         Di~Kang, 
         Tingguang~Li, 
        Ruizhi~Chen, 
        and~Linchao~Bao \IEEEauthorrefmark{1}
\IEEEcompsocitemizethanks{
\IEEEcompsocthanksitem W. Wang and Q. Ke are with the Faculty of Information Technology, Monash University, Melbourne 3168, Australia. 
\protect\\ 
E-mail: weiqking@gmail.com, qiuhong.ke@monash.edu.

\IEEEcompsocthanksitem X. Zhe, D. Kang and L. Bao are with Tencent AI Lab, Shenzhen 518054, China. 
\protect\\ 
E-mail: \{zhexuefei,~di.kang\}@outlook.com, linchaobao@gmail.com.

\IEEEcompsocthanksitem T. Li is with Tencent Robotics X, Shenzhen 518054, China. 
\protect\\ 
E-mail: tgli0809@gmail.com.

\IEEEcompsocthanksitem
R. Chen is with the State Key Laboratory of Information Engineering in Surveying, Mapping, and Remote Sensing, Wuhan University, Wuhan 430079, China. 
\protect\\ 
E-mail: ruizhi.chen@whu.edu.cn.

\IEEEcompsocthanksitem W. Wang did the work during his internship at Tencent AI Lab.\\
 \IEEEauthorblockA{\IEEEauthorrefmark{2} W. Wang and X. Zhe contributed equally to this work. }\\
 \IEEEauthorblockA{\IEEEauthorrefmark{1} L. Bao is the corresponding author. }%

}
}

\IEEEtitleabstractindextext{%
\begin{abstract}
We present a neural network-based system for long-term, \emph{multi-action} human motion synthesis. 
The system, dubbed as {\systemname}, can produce high-quality and meaningful motions with smooth transitions from simple user input, including a sequence of action tags with expected action duration, and optionally a hand-drawn moving trajectory if the user specifies.
The core of our system is a novel Transformer-based motion generation model, namely {\thename}, which can generate diverse motions given action tags. 
Different from existing motion generation models, {\thename} utilizes contextual information from the past motion clip and future action tag, dedicated to generating actions that can smoothly blend historical and future actions. 
Specifically, {\thename} first encodes the target action tag and contextual information into a latent code at the action level. 
The code is unfolded into frame-level control signals via a time-unrolling module, which could then be combined with other frame-level control signals, such as the target trajectory. 
Motion frames are then generated in an auto-regressive way. 
By sequentially applying {\thename}, the system {\systemname} can robustly generate long-term, multi-action motions with the help of two simple schemes, namely ``Shadow Start'' and ``Action Revision''. 
Along with the novel system, we also present a new dataset dedicated to the multi-action motion synthesis task, which contains both action tags and their contextual information. 
Extensive experiments are conducted to study the action accuracy, naturalism, and transition smoothness of the motions generated by our system. Project page: \url{https://wjohnnyw.github.io/blog/tag2motion/}
\end{abstract}

}

\maketitle

\IEEEdisplaynontitleabstractindextext

%
\IEEEpeerreviewmaketitle

\begin{figure*}[!t]
    \centering
    \begin{overpic}[trim=2cm 0.9cm 2cm 1cm,clip,scale=0.5,height=0.4\linewidth,width=0.95\linewidth,grid=false]{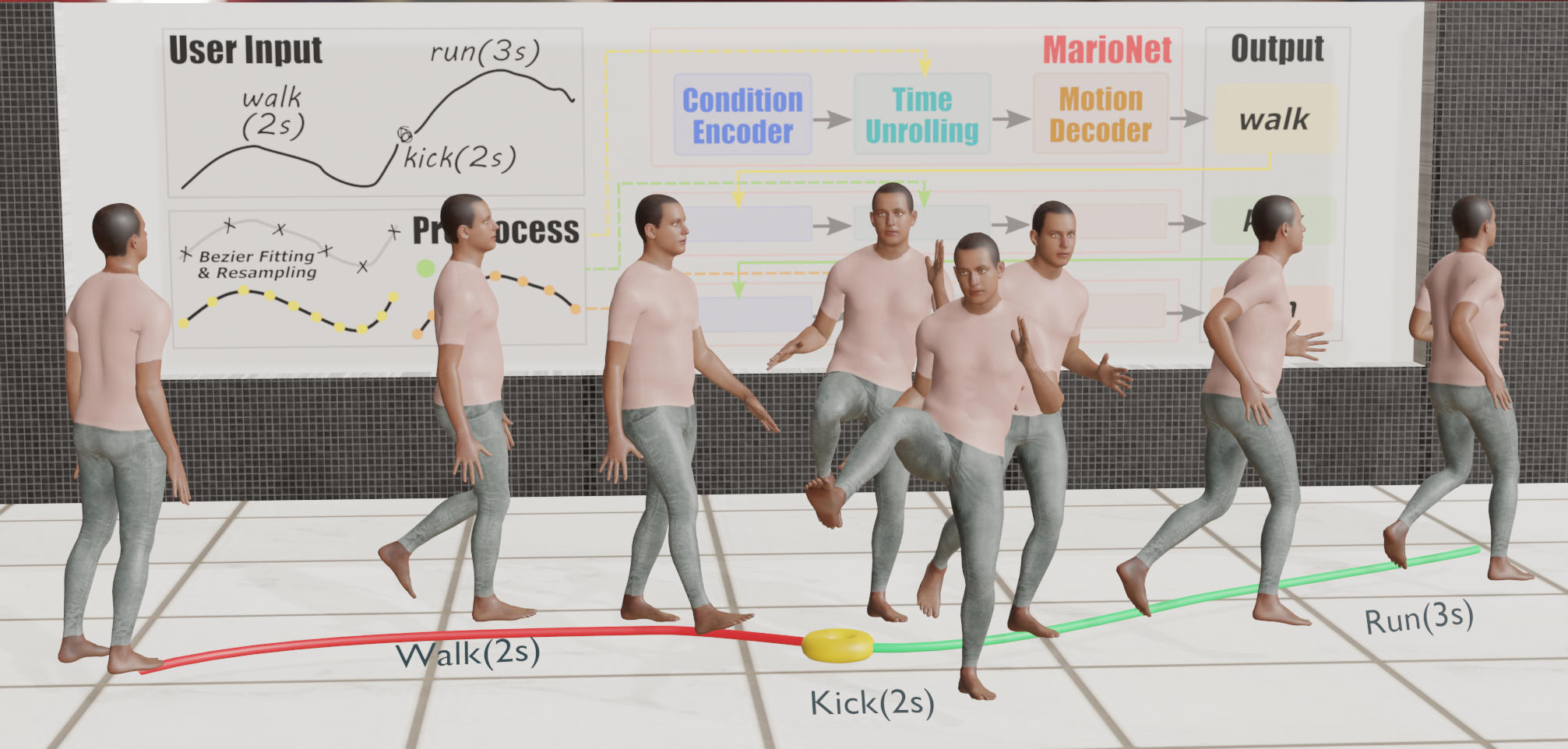}
    \end{overpic}\vspace{-8pt}
    \caption{
    Our system {\systemname} only needs a user to draw a trajectory line with the action tag for each segment and can produce a 
    plausible motions with smooth transitions between actions. The core of the system is a novel Transformer-based neural network named {\thename}.
    \vspace{-15pt}
    }
    \label{fig:eg_run}
\end{figure*}

{\section{Introduction}
\label{sec:introduction}}

\IEEEPARstart{H}{uman} motion capturing, modeling, and synthesis have been widely studied for decades, due to the rapidly increasing demand from the entertainment industry. 
Although new motion capture equipment and the latest motion processing algorithms have made motion collecting more accurate and efficient, it is still a challenging task to synthesize meaningful long-term human motions in a controllable yet simple way.
%

We observe that a meaningful long-term human motion can be easily described by a sequence of semantic action tags. Thus, one naturally raises the following question: Can we use a sequence of action tags together with corresponding action duration as control signals to generate a realistic and smooth human motion clip?

Benefiting from the rapid development of deep generative models, recently proposed deep neural network-based methods have demonstrated promising achievements in motion synthesis.
%
Animation synthesis for game control is well studied in \cite{PFNN,zhang2018mann,holden2020learned}, which has shown superior quality in the locomotion synthesis task.
However, most locomotion synthesis methods are tailored to the gaming industry with limited model size to reduce response time to improve players' experience, which constrains their ability to generate diverse and complicated motions.
Moreover, these locomotion synthesis methods generate motion in a frame-by-frame scheme without having an overall scope
of the complete action, which is another prominent limitation for producing non-periodical actions such as "kick" or "throw". 
On the other hand, two recent works, action2motion~\cite{guo2020action2motion} and ACTOR \cite{actor}, can generate diverse motions from a set of semantic action tags.
However, both of them are elaborately designed to be conditioned on a single action tag, thus hard to incorporate other contextual information, such as initial poses for a more controllable generation. 
Other works \cite{yuan2020dlow_pred,barsoum2018hp_pred,habibie2017recurrent,wang2019combining,wang2021spa} can generate continuous motion based on previous poses.
However, these methods are only suitable for short-term motion prediction and cannot be used to generate a preferred action from a specified category.

We can see that existing motion synthesis methods as mentioned above still struggle to generate meaningful long-term motions from a set of action tags. More specifically, there are three major challenges:
(1) The first challenge is that there is a lack of suitable datasets for this task.
%
Early datasets, such as \cite{guo2020action2motion,shahroudy2016ntu,liu2019ntu} only provide action tag and motion data without any contextual information, which makes it hard to generate continuous actions.
The latest motion dataset named BABEL~\cite{babel} provides frame-level motion annotations, which can be considered as containing contextual information such as previous motion and future tag.
However, due to the lack of a clear definition of each action, the annotated tags in BABEL are very noisy and ambiguous, making it hard to train a motion generative model.
(2) The second challenge is how to formulate and encode contextual information into the long-term motion generation task.
In practice, we have found it challenging to train a model which \emph{directly} generates a long-term motion only with sequential action tags. 
One optimal solution is to decompose this complicated action tag sequence into motion sequence generative problem into single action generation blocks, where each action is generated with a smooth transition to the previous one.
Then the expected long-term motion can be obtained by stitching these continuous actions together.
In other words, a powerful generative model that takes not only the current semantic action tag but also contextual information is required. 
Unfortunately, there are few existing methods that can address the requirements.
(3) The third challenge is that errors or unnaturalness in synthesized motions can accumulate quickly over time.
Each separate action in multi-action generation needs additional contextual information to ensure a smooth transition between actions, compared to the original single-action generation which only considers target tag information. 
However, the contextual information that comes from the previously generated motion may not always be in the distribution seen in the training. Thus, errors would gradually pass through the sequential process of multiple action generation if proper correction strategies are not adopted. 
It is quite similar to the common drifting problem in a temporal prediction model.

To address these three challenges, we propose our solution from three aspects.
(1) First, we present a new BABEL-based dataset, named \emph{ BABEL for Multi-Action Generation} ({\dtname}).
We collect motion clips from the first 40 largest action categories from BABEL. 
Then we recategorize them into 31 action groups following a well-defined and consistent labeling protocol.
We also provide \emph{contextual information} for each action, including initial motion and the tag of the next action, together with the current action tag.
(2) We propose a single-action generative model based on Transformer ~\cite{vaswani2017Transformer} that takes both the current action tag and contextual information as input.
%
Specifically, our model first encodes the target action tag and contextual information into an action-level latent code $Z$.
Then this action-level latent code unfolds into time-varying frame-level codes with our time-unrolling net.
The frame-level codes provide better duration control and more stable single-action generation. 
Besides, other control signals such as trajectory can be combined together with frame-level codes during this process.
Finally, the entire action can be obtained frame by frame with our auto-regressive motion decoder.
With the help of contextual information, actions that are generated in sequential order can be stitched easily to produce a long-term motion with smooth transitions.
(3) Lastly, to alleviate error accumulation, we propose a recognition-guaranteed hybrid pipeline to generate a continuous motion with multiple actions, consisting of a feature projection scheme named \emph{``Shadow Start''} and a recognition-based scheme named \emph{``Action Revision''}. 
As a result, we propose an effective and easy-to-use motion synthesis system that can generate accurate, natural human motions with smooth transitions from a sequence of action tags.
Extensive experiments on both single- and multi-action generation are presented to demonstrate the superiority of our proposed method.

\textbf{Contributions.} Our contributions can be summarized as follows:
\begin{itemize}
[leftmargin=*]
    \item We propose a novel single-action generation network, namely {\thename}, which utilizes both semantic tag and contextual information, to generate transition-smooth actions. Furthermore, {\thename} surpasses state-of-the-art single-action generative models in terms of action accuracy and naturalism.
    \item Based on {\thename}, we present a system {\systemname} to iteratively predict long-term multiaction human motion. Two robust schemes, called \emph{``Shadow Start''} and \emph{``Action Revision''} are introduced to deal with the error accumulation problem, which significantly improves the capability of the system for generating long-term motion sequences. 
    \item We introduce a new motion dataset {\dtname} for the multi-action motion generation task, which provides both action tags and their contextual information. 
\end{itemize}

Our paper is organized as follows. We first review some motion synthesis literature in Sec.~\ref{sec:relatedWork}. In Sec.~\ref{sec:dataset}, we analyze the limitations of existing motion datasets and propose our new {\dtname} dataset, which is more suitable for multi-action synthesis.
We then introduce {\thename} with full details in Sec.~\ref{sec:mtd}, which is the key building block of our {\systemname} system in the following subsection.
Extensive experiments on different aspects are presented in Sec.~\ref{sec:experiment} and we demonstrate how to efficiently generate desirable motions with an interactive GUI in Sec.~\ref{sec:system}.
Finally, we summarize the advantages, limitations and future work of our method for motion generation.




{\section{Related Work}}
\label{sec:relatedWork}

\subsection{Action-Conditioned Motion Synthesis}
Action-conditioned human motion synthesis aims at generating realistic and diverse human motions corresponding to specified action tags.
This is an under-explored field compared to other well-defined tasks including locomotion generation~\cite{holden2016deep,habibie2017recurrent,PFNN,zhang2018mann}, dance synthesis~\cite{tang2018dance,lee2019dancing,li2020learning,li2021ai,Aristidou2022Rhythm,chen2021a}, and gesture synthesis~\cite{levine2010gesture,ginosar2019learning,yoon2020speech,li2021audio2gestures}.
Earlier works~\cite{guo2020action2motion,actor} trained on exiting motion dataset~\cite{shahroudy2016ntu,liu2019ntu,ji2018large} are able to generate different types of actions,
including locomotion, exercises, and some daily actions.
%
However, \cite{guo2020action2motion,actor} cannot be applied without modifications to generate a relatively complicated motion consisting of a sequence of different actions. 
Blending multiple independently generated actions cannot guarantee natural transitions and smooth motion trajectories.
In order to generate consecutive actions with natural transitions, besides the action tag, our network is additionally conditioned on contextual information for motion synthesis.
%

%

\subsection{Control Signals in Motion Synthesis}
There are several well-known motion synthesis subfields categorized according to the motion types, including locomotion synthesis, dance motion synthesis, co-speech gesture synthesis, etc. 
Here we focus on the \emph{control signals} used in various motion synthesis tasks.

Some literature take \emph{dense} control signal for every \emph{frame} as input.
For example, the locomotion synthesis task usually specifies dense signals for motion trajectory including turning, forward, and sideways velocity for every frame \cite{habibie2017recurrent,wen2021autoregressive}.
Later, other control signals including foot contact~\cite{holden2016deep}, phase~\cite{PFNN}, or pace~\cite{pavllo2018quaternet} are utilized to resolve motion ambiguity during the prediction, achieving better results. 
Dense control signals could help the network avoid generating static poses or floating characters~\cite{habibie2017recurrent,holden2016deep,henter2020moglow} in locomotion, which are major challenges for all types of long-term motion synthesis tasks.
Similarly, in the dance motion synthesis task, dense acoustic features extracted from the input music such as beat, onset strength, MFCC and etc. are used to guide the motion generation~\cite{lee2019dancing,tang2018dance,huang2020dance,dance2beat}.
Most previous works~\cite{lee2019dancing,tang2018dance,zhuang2020music2dance,huang2020dance} only take music (features) as input to generate dance motions, formulating it as a cross-modal translation task.
A recent paper~\cite{li2021ai} takes an additional 2-second seed motion along with the music features as input to generate the following dancing poses, which can be equivalently regarded as a motion prediction task conditioned on input music.
\cite{li2021ai} can generate better dance motions over the state-of-the-art methods that only take music as input since the presence of the seed motion significantly reduces the possible motion space for the following motions.
We can observe that introducing proper dense and informative control signals can potentially improve the results of motion synthesis.
Inspired by this observation, we utilize previous poses as additional control signals so that the generated actions can be concatenated to accomplish a complex task including a series of actions.

Some other works take \emph{sparse} control signal that stays the same for each frame in one action.
For example, \cite{guo2020action2motion,actor} take as input the target action tag and generate diverse motions as expected.
This setting with limited input information is severely challenging since adapting dense-signal conditioned methods aforementioned to handle the sparse control signal will quickly result in generating static poses as demonstrated in \cite{guo2020action2motion}. 
In order to alleviate this difficulty and generate diverse motions belonging to the same category, \cite{guo2020action2motion,actor} adopt a probabilistic VAE framework that can take a random sample as input during inference.
However, these methods can only synthesize a single action at one time and cannot generate long and consecutive motions like locomotion, dance, and co-speech motion synthesis methods~\cite{PFNN,ginosar2019learning,lee2019dancing,li2021audio2gestures}.

\subsection{Deep Learning in Motion Synthesis}
Traditional motion synthesis methods usually rely on a search-and-blend framework to \textit{concatenate} short motion clips from an existing motion database. 
Representative methods include motion graph-based methods~\cite{kovar2002motion}, which formulate motion synthesis as a path searching problem in a graph constructed based on motion similarity for all the motion clips in the database.
Various constraints can be incorporated into the path searching process, making this framework very flexible. 
For example, a recent graph-based method~\cite{chen2021choreomaster} has incorporated deeply learned high-level style and rhythm features of choreography-music to search for the most suitable candidates, achieving convincing dance choreography.

However, relying on neural networks to directly synthesize motion instead of searching-and-blending short motions from an existing database is becoming more and more popular since network-based methods can encode database information into their weights instead of storing a huge motion database~\cite{holden2020learned} and it is possible to generate novel and realistic motions that have not been seen in the training set~\cite{actor}.
For example, early deep learning methods mostly adopt the deterministic framework for motion synthesis. 
\cite{fragkiadaki2015recurrent} proposed a representative Encoder-Recurrent-Decoder (ERD) framework to jointly learn spatial representations (with the encoder and decoder) and their dynamics (with a recurrent 
network) to predict the next frame pose. 
\cite{martinez2017human} predicts both short-term and long-term future motion by GRU layers~\cite{cho2014properties}  with a novel sampling-based loss.
\cite{holden2016deep} uses the MLPs for mapping the control signal to a hidden motion space which is later modelled by the CNN autoencoder to generate corresponding locomotion.

Recently, probabilistic framework gains more and more attention since it can better avoid generating frozen motion and generating diverse results by sampling~\cite{habibie2017recurrent,lee2019dancing,henter2020moglow,li2021audio2gestures,guo2020action2motion,actor}. 
\cite{lee2019dancing} proposes a dance unit VAE to encode and decode the dance units and a music-to-movement GAN to generate dance movement conditioned on the input music.
\cite{li2021audio2gestures} utilizes a TCN-VAE framework and split the latent space into two parts to explicitly model the one-to-many mapping between speech and gesture, producing more diverse and realistic results. 
\cite{henter2020moglow} proposes to model the conditional probability distribution of the next frame pose using normalizing flows that takes previous poses and control signals as input.
\cite{guo2020action2motion} proposes a GRU-VAE and \cite{actor} proposes a Transformer-VAE to generate realistic and diverse motions belonging to the given action category.
We also adopt a Transformer-based probabilistic framework but introduce several key designs in order to synthesize complex motions containing multiple actions.

\section{{\dtname} Dataset}\label{sec:dataset}

To synthesize high-quality and continual human actions, we need a large-scale tag-labelled motion dataset.
A good dataset for continuous action synthesis should satisfy the following criteria: 
(1) The motion sequences in the dataset should contain multiple actions that can provide contextual information for continual action generation.
(2) The action tags should be \emph{precise and unambiguous}.
For example, the same tag should not be used to describe two actions with different movement patterns. 

\begin{figure}[!t]
    \centering
    \begin{overpic}[trim=0cm 0cm 0cm 0cm,clip,width=1\linewidth,grid=false]{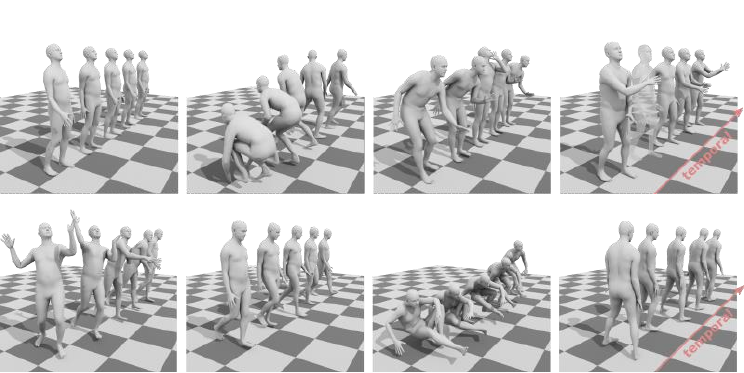}
    \put(0,46.5){\scriptsize \textbf{"standing"}}
    \put(25,46.5){\scriptsize \textbf{"take sth. up"}}
    \put(50,46.5){\scriptsize \textbf{"using object"}}
    \put(75,47){\scriptsize \textbf{"arm movements"}}
    \put(75,45){\scriptsize \textbf{"standing"}}
    
    \put(0,21.5){\scriptsize \textbf{"throw"}}
    \put(25,21.5){\scriptsize \textbf{"walk"}}
    \put(75,21.5){\scriptsize \textbf{"step"}}
    \put(50,21.5){\scriptsize \textbf{"stand up"}}
    
    \end{overpic}\vspace{-12pt }
    \caption{\textbf{Labeling error in BABEL}. Descriptions of BABEL can be ambiguous. For example, the upper actions miss the most prominent movement and are labeled as (1) "arm movements\&raising body part", (2) "turn\&head movements", (3) "hand movements", and (4) "stand" in BABEL respectively while lower actions are all labeled as "stand" despite of significantly different movements. For comparison, our tagging rules for {\dtname} can easily separate different types of actions and the corrected tags are annotated above the images.}\vspace{-10pt}
    \label{fig:dt:babel:err}
\end{figure}

Existing datasets such as~\cite{guo2020action2motion,liu2019ntu,ji2018large} can only support action-level synthesis (i.e., synthesize a motion with respect to a single tag one time). 
A recent dataset, BABEL~\cite{babel}, provides motion annotations to the large-scale MoCap dataset AMASS~\cite{amass}.
Its motion annotation includes frame-level tags in continual motion sequences, which can be potentially used for multi-action synthesis. 
However, the action tags in BABEL do not fully satisfy the criterion (2), which poses difficulties in the training motion generative models.
As we observe (also mentioned in the original paper~\cite{babel}), BABEL has some significant issues: 
(1) The action tags are \emph{unbalanced} and have a long-tailed distribution, which makes it difficult for network training.
(2) Since the action tags are abstracted from sentences proposed by different annotators, some action tags are \emph{ambiguous}, which can hardly be used for motion synthesis. For example, the tag "turn" is used for describing a body turn-around, walking in circles, and even head-turning; the tag "step" is used for both strolling and climbing stairs.
(3) The action tags can be composed but no clear \emph{composition rules} are discussed or presented, which makes it hard for synthesizing diverse and complicated actions in real life. See Fig.~\ref{fig:dt:babel:err} for some qualitative illustrations.

To avoid these issues above, we present a new dataset, {\dtname}, by relabeling actions from the first $40$ \emph{largest} and as \emph{meaningful} as possible categories in BABEL, containing over 3 hour motion data for ''Multi-Action Generation'' task.
First of all, we abstract \boldsymbol{$31$} action tags from the 40 BABEL motion category names, including \boldsymbol{$24$} tags for describing action intentions, \boldsymbol{$3$} for body states, \boldsymbol{$4$} for body part movements.
For intention-describing tags, we further divided them into \emph{root- motion} tags and \emph{in-place} tags, depending on whether the common trajectories of these motions change significantly.
As a result, we successfully resolve the tagging ambiguity in the chosen \boldsymbol{$7643$} actions from BABEL. See Fig.~\ref{fig:dt:babel:err} for some examples with corrected tags in {\dtname}. 
In addition to action tags, we also provide initial motion and future action tags describing the original following motion as action contextual information. That is, for every item, it includes: (1) $k$ frames of previous motion ($\{P\}_{-k}^{-1}$)); (2) current action tag ($A^{cur}$); (3) current action duration ($T$); (4) next action tag ($A^{next}$). And each motion ${P}$ is represented as a joint rotation with root translation per frame as ~\cite{amass,SMPL-X:2019}.
See Sect. 2 in the Supplementary for more details about {\dtname} labeling. 
\section{Methodology}
\subsection{Problem formulation}
Given a sequence of $n$ action tag annotations $\Scale[0.95]{\mybrac{A_i}_{i=1\cdots n}}$  with corresponding expected action duration $\Scale[0.95]{\mybrac{T_i}_{i=1\cdots n}}$ and optionally user defined trajectories $\Scale[0.95]{\mybrac{U_i}_{i=1\cdots n}}$ (we set $U_i$ to be zero-value if not specified), 
our goal is to generate a \emph{sequence of actions} $\Scale[0.95]{\mybrac{M_i}_{i=1\cdots n}}$ where the action $\Scale[0.95]{M_i = \mybrac{P_1, \cdots, P_{T_i}}}$ can be accurately described by the input label $A_i$, and follow the input trajectory $U_i$.
%
More importantly, the motion sequence is expected to be realistic with smooth transitions between any two actions $M_{i}$ and $M_{i+1}$. 

However, it is challenging to directly synthesize long-term human motion with deep generative models. 
Instead, we first propose a single action generative model, named as~\thename, then demonstrate a hybrid system  named as {~\systemname} synthesize multi-action motion by sequentially using ~\thename.

\subsection{{\thename}: Single-action Generation Model}
\label{sec:mtd}
\begin{figure}[!h]
    \centering
    \includegraphics[width=1\linewidth]{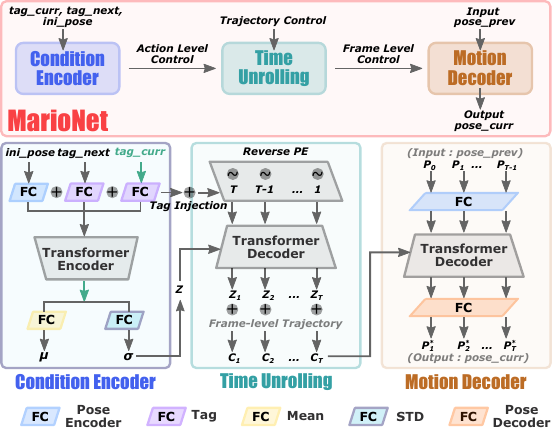}\vspace{-5pt}
    \caption{\textbf{Network Overview of {\thename}}.  {\thename} contains three major building blocks: Condition Encoder block, Time Unrolling block, and Motion Decoder block. The fully connected (FC) layers with the same color share the same weights.}
    \label{fig:mtd:network}
\end{figure}

In this part, we discuss the design details of {\thename}, which is the main building block of our motion sequence generating pipeline (illustrated in Fig.~\ref{fig:mtd:network}). 
{\thename} can synthesize a single action that accurately corresponds to the input action tag and smoothly transits from the initial motion while taking the future action tag into consideration.

\subsubsection{Network structure}
{\thename} is an auto-regressive attention-based model \emph{conditioned} on (1) initial motion $P^{\text{ini}}$, (2) current action tag $A^{\text{curr}}$, and (3) the following action tag $A^{\text{next}}$, which contains three main building blocks with Transformer layers, namely the Condition Encoder, Time Unrolling Module, and Motion Decoder.


\emph{Condition Encoder.}
The Condition Encoder projects inputs (initial poses, current and next action tags) to an \emph{action-level} latent distribution parameterized by mean $\mu$ and variance $\sigma$.
%
Specifically, we first embed pose and tag information to feature space with different FC layers. Then embedded $P^{\text{ini}}, A^{\text{curr}}, A^{\text{next}}$ are concatenated together and fed into a Transformer encoder.
We are particularly interested in the embedded $A^{\text{curr}}$ (dubbed as \emph{action-token}). 
Similar to ACTOR~\cite{actor}, only the action token is used for pooling. 
The output of the Transformer encoder that corresponds to the action token is used to regress the distribution parameters $(\mu, \sigma)$ via two independent FC layers.
Sampling from the obtained distribution leads to an \emph{action-level} latent code $Z$.

\emph{Time Unrolling Module.} 
This module learns a \emph{frame-level} latent space by unrolling the action-level latent code and potentially combining user frame-level input information (e.g. trajectory), which we call frame-level controls as a whole. 
Specifically, the \emph{action-level} latent code $Z$ from previous step is firstly unrolled into \emph{frame-level} latent codes $[z_1, \cdots, z_T]$ via a classic Transformer Decoder. 
The Transformer Decoder takes the action-level latent code $Z$ as Key $K$ and Value $V$, and takes the \emph{reverse} positional encoding (Rev-PE) with \emph{Tag Injection} operation (described in the next paragraph) as Query $Q$ for cross-attention.
As a result, we get the corresponding frame-level latent codes $[z_1, \cdots, z_T]$ in $\mathbf{R}^{d}$.
At this stage, we can also append other frame-level information such as motion trajectories to better guide or control the motion synthesis. 
To achieve this, the latent codes $[z_1, \cdots, z_T]$ is first projected into $\mathbf{R}^{\tilde{d}}$ via a FC layer, while the user input is projected into $\mathbf{R}^{d - \tilde{d}}$ via another FC layer.
Combining them together leads to the finalized frame-level controls  $[c_1, \cdots, c_T]$ in $\mathbf{R}^{d}$.
In this way, the frame-level trajectory can be concatenated with frame-level action latent space, resulting in more powerful frame-level controls.
We adopt Rev-PE to encode positions for time unrolling inspired by the time-to-arrival (TTA) embeddings in~\cite{harvey2020robust}, which is beneficial for motion generation with a pre-defined number of frames since the action termination information is encoded in the Rev-PE. 

\emph{Tag Injection.}
Due to different signals being mixed up in the condition encoder for an action-level representation, the control ability of the current action tag $A^{\text{curr}}$ which we expect to be dominant is inevitably weakened. 
Therefore, we propose a tag injection operation to enhance its control ability, leading to generated action more exactly corresponding to the specified tag.
Specifically, we add the \emph{embedded} $A^{\text{curr}}$ to every frame of Rev-PE, then use them as the Query $Q$ in Time Unroll Module for cross-attention.

\emph{Motion Decoder.}
The Motion Decoder takes a sequence of poses $P = [P_0, \cdots, P_{T-1}]$, and the frame-level controls $C = [c_1, \cdots, c_T]$ as input, and outputs the synthesized action $P^* = [P^*_1, \cdots, P^*_T]$, i.e., a set of new poses shifted 1 frame right temporally from the input poses.
Specifically, the input poses $P_t$ are firstly projected into the same pose latent space as the one discussed in the Condition Encoder via an FC layer.
Then the Transformer Decoder takes the embedded input poses with positional encodings as Query $Q$, and takes the frame-level controls $c_t$ as Key $K$ and Value $V$ for cross-attention.
The outputs of the Transformer Decoder are then recovered into pose space by linear projections with FC.
During the training phase, the Transformer Decoder with attention-mask can process the ground-truth input poses in parallel. 
During the test, the complete action with $T$ frames is generated iteratively. Specifically, at each iteration, one new frame is firstly generated and then feed to the Transformer Decoder again to synthesize a new pose in the following frame.

\subsubsection{Training}
In this part, we discuss in detail how we train {\thename} on {\dtname} and how we construct the training loss for the motion synthesis task.

%
\begin{figure}[!h]
    \centering
    \begin{overpic}[trim=0cm 27.2cm 14.1cm 0cm,clip,width=0.9\linewidth,grid=false]{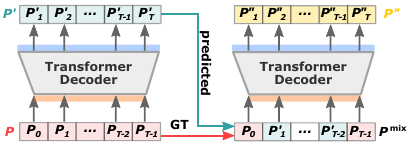}
    \end{overpic}\vspace{-12pt}
    \caption{\textbf{Scheduled Sampling Strategy}. \emph{Left}: the Motion Decoder takes the ground-truth poses $\{P_t\}_{t=0\cdots T-1}$ as input and output the synthesized poses $\{P'_{t}\}_{t=1\cdots T}$. \emph{Right}: $\{P_t\}$ and $\{P'_t\}$ are \emph{mixed} and feed into the Motion Decoder again to synthesize new poses $\{P''_t\}_{t=1\cdots T}$. Both $P'_t$ and $P''_t$ are used for loss construction. For simplicity, we use an orange/blue box to represent the pose encoding/decoding FC layer.  
    } \vspace{-10pt}
    \label{fig:mtd:sampling}
\end{figure}
 \emph{Scheduled Sampling for Transformer.} To alleviate this dilemma in sequence generation training, we adapt the Scheduled Sampling~\cite{bengio2015scheduled,li2021scheduled} to motion generation with Transformer.
As illustrated in Fig.~\ref{fig:mtd:sampling}, two weights-sharing Motion Decoders with Transformer Decoder are involved. More details can be found in the supplementary.

\emph{Training Loss.} We use the following loss to measure the reconstruction quality:
\begin{subequations}\label{eq:mtd:recon_loss}
\begin{align}
    \Scale[1.0]{L_{\text{Recon}}\big(\hat{P}\big) } & \, \Scale[0.95]{= \lambda_R L_R + \lambda_D L_D + \lambda_J L_J},\\
    \text{where} \quad \Scale[0.75]{ L_R } & \,\Scale[0.75]{ = \frac{1}{T} \sum\limits_{t = 1}^{T} \big\Vert\, R_t - \hat{R}_t \,\big\Vert_2^2 + \lambda_V \frac{1}{T-1}\sum\limits_{t=2}^{T} \big\Vert\, R'_{t} - \hat{R}'_{t}  \,\big\Vert_2^2},\\
    \quad \Scale[0.75]{ L_D } & \, \Scale[0.75]{ = \frac{1}{T} \sum\limits_{t = 1}^{T} \big\Vert\, D_t - \hat{D}_t \,\big\Vert_2^2  + \lambda_V \frac{1}{T-1}\sum\limits_{t=2}^{T} \big\Vert\, D'_{t} - \hat{D}'_{t}  \,\big\Vert_2^2},\\
    \quad \Scale[0.75]{ L_J } & \, \Scale[0.75]{ = \frac{1}{T} \sum\limits_{t = 1}^{T} \big\Vert\, J_t - \hat{J}_t \,\big\Vert_2^2  + \lambda_V \frac{1}{T-1}\sum\limits_{t=2}^{T} \big\Vert\, J'_{t} - \hat{J}'_{t}  \,\big\Vert_2^2}
\end{align}
\end{subequations}
where the rotation $\hat{R}_t$, root translation $\hat{D}_t$, and joint position $\hat{J}_t$ are associated with the synthesized pose $\hat{P}_t$; $\hat{R}'_{t} = \hat{R}_{t} - \hat{R}_{t-1}$ give the rotation velocity of that pose (similar for the root translation and joint position).
Specifically, for a given predicted pose $\hat{P}_t$, which is represented by the rotations $\hat{R}_t$ and a root translation $\hat{D}_t$, we can then fit a SMPL model~\cite{smpl} with mean shape parameters to get the 3D positions for the joints $\hat{J}_t$ of the pose $\hat{P}_t$. After this, we can then measure the reconstruction error of the motion sequence $\hat{P}$ by computing the per-frame differences and velocity difference of rotation, root translation, and joint position between the predicted motion and the ground-truth motion.
Therefore, we have the \emph{Reconstruction Loss on Rotation} $L_R$ defined in (Eq. (\ref{eq:mtd:recon_loss}b)); the \emph{Reconstruction Loss on Translation} $L_D$ defined in (Eq. (\ref{eq:mtd:recon_loss}c)); and the \emph{Reconstruction Loss on Local Joint Positions} $L_J$ defined in (Eq. (\ref{eq:mtd:recon_loss}d)).  
We also adopt the Kullback-Leibler (KL) divergence to regularize the distribution of the action-level embedding in our Condition Encoder. 
In summary, we have the following loss function for training:
\begin{equation}
\Scale[0.95]{ L = \lambda_1 L_{\text{Recon}}\big(P'\big) + \lambda_2 L_{\text{Recon}}\big(P''\big) +   \lambda_{\text{KL}} L_{\text{KL}} }
\end{equation}
Note that, $\lambda_R, \lambda_D, \lambda_J, \lambda_V, \lambda_1, \lambda_2$ are weights used for balancing the realism and diversity by manual setting. 

\subsubsection{Inference}
\label{sec:inference}
During the inference phase of generating action with tag $A$, the current tag $A^{\text{curr}}$ (set as $A$), the initial motion $P^{\text{ini}}$ that corresponds to tag $A$, and the following tag $A^{\text{next}}$ are firstly fed into the Condition Encoder and then $\mu$ and $\sigma$ are obtained.
An action-level latent code is sampled from $\mathcal{N}(\mu,\,\sigma^{2})$ then combined with $T$ frames reverse PE (corresponding to specified action duration), and trajectory $U$ constraints (set to zero-value for in-place actions) to generate the frame-level controls.
Finally, the Motion Decoder runs in a frame-by-frame fashion to auto-regressively generate the motion sequence that corresponds to the expected action tag $A$ from the frame-level controls. 
%
%
For the first iteration, we use the last frame of $P^{\text{ini}}$ as the Begin of Poses (BOP). We terminate this process when we obtain $T$ frames of predicted poses as requested. 
\subsection{Generating Multi-action Motion}
Previous parts introduce how to generate a single action conditioned on both tag and contextual information by ~\thename.
In this section, we discuss how to synthesize a long-term multi-action motion by sequentially applying the {\thename}.  
We first introduce a pure generative pipeline, which only relies on the generative model such as {\thename}.
Then based on this pipeline, we derived ~\systemname with two simple but effective correct schemes for a more robust multi-action generation.
\subsubsection{Pure generative pipeline}
A simple way to synthesize multi-action motion is to generate multiple single-action motions with an action generative model, then connect these singe-action clips temporally.
%
%
Specifically, we first normalize initial motion ($x$ frames) $P^{\text{ini}}=\{P_t\}_{t=-x}^{t=-1}$  to facing $-y$ direction (zero-rotation of T-pose) then moved to origin point both based on its ending pose $P_{-1}$, where pose $P_{-1}$, which is the last frame pose of previously generated action or a given initial motion, has a global rotation $R_z = (0,0,\gamma)$ on z-axis, ($R = (\alpha,\beta,\gamma)$ in x-y-z form of Euler angle) and a global translation $D_t(x,y,z)$.
For each frame of normalized initial motion:
\begin{equation}
    \{\, P^*_t\,\}_{t=-x}^{t=-1} = R_z^{-1} \{\, P_t\,\}_{t=-x}^{t=-1} - D_t
\end{equation}
After obtaining the normalized initial motion, an action is generated by applying the {\thename} as described in Sec.~\ref{sec:inference}:
\begin{equation}
\label{formula:marionet}
\{\tilde{P_t}\}_{t=1}^{t=T} = {\thename}\big( \{\, P^*_t\,\}_{t=-x}^{t=-1},  A^{\text{curr}},A^{\text{next}},T,U\big)
\end{equation}
For stitching with the previous action, the generated action needs to be rotated back by 
\begin{equation}
\{\,\hat{P_t}\,\}_{t=1}^{t=T} = R_z \{\tilde{P_t}\}_{t=1}^{t=T} + D_t 
\end{equation}
Then the following actions can be generated by sequentially applying these steps.
%
%

%
\subsubsection{{\systemname}}\label{subsec:hybrid}

%
During testing, we notice that the pure generative pipeline may fail sometime, especially when generating long multi-action motions, i.e., motions including more than $10$ actions. 
Through further digging into this problem, we find two main reasons.
The first reason is \emph{insufficient} connected pairs of actions for training.
%
%
By counting connection actions tags, we find that, except for a few tag pairs (e.g. turn-walk), most of the tag pairs have only limited connected samples in the training data. (See more detail in Supp. Fig 1.)
A single-action network trained on data with such a sparse connection matrix tends to face a severe out-of-distribution problem when generating continuous actions.
A straightforward solution is to obtain more connected motion capture data.
However, it is not feasible because the desirable data grows with the tag number by the magnitude of the square.
The second reason is the error accumulation issue, which is a major challenge faced by many temporal generative models when generating long sequences.
To tackle the above challenges, we propose a hybrid system named ``\systemname'', which includes two important parts: ``Shadow Start'' and ``Action Revision''.

\emph{Projection based ``Shadow Start'' }
Inspired by solutions for long-term tracking~\cite{kalal2011tracking,Zhu_2018_ECCV_tracking} and the manifold projection in DeepFaceDrawing~\cite{chen2020deepfacedrawing}, we proposed a pose embedding projection method to improve the stability of long-term generation, named ``Shadow Start''.
The main idea of Shadow Start is to project the feature embedding of Begin of Pose (BOP) $P_0$, as mentioned in Sec.~\ref{sec:inference}, to a hyper-plane supported by some embeddings derived from BOPs in the training set.
More specifically, the BOP embedding feature, $f_0$ is firstly extracted by $f_0 =FC(P_0)$, where $FC$ is the pose embedding layer in Fig.~\ref{fig:mtd:network}. 
Then, $N$ most similar embeddings extracted from BOPs $\{\tilde{P}^i\}_N$ belonging to the current action tag $A^{\text{curr}}$ in the training dataset are retrieved according to their $L2$ distance to $f_0$ in the embedding space.
Here, pose embeddings of $\{\tilde{P}^i\}_N$ are denoted as $\tilde{F} = [\tilde{f}^i]_N$.
Following this, the pose embedding $f^*_0$ of desired projected $P_0^*$ can be found by solving the linear combination coefficients $ \Phi= [\phi_i]_N$ as follows:
\begin{equation}
      \Scale[0.95]{{\Phi}^* = \text{argmin}_\Phi ||f_0  - \Phi \cdot \tilde{F} ||_2^2},~~~f^*_0 =  \Phi^* \cdot \tilde{F}
\end{equation}
%
Since this simple projection can not guarantee whether $f_0^*$ is on the meaningful region.%
We further use a $L2-$normalization to $f_0^*$, which help us project pose to a compact hyper-sphere space.
After obtaining $f_0^*$, the new action can be generated as described in Sec.~\ref{sec:inference} only by replacing the origin, BOP embedding $f_0$ with $f_0^*$.

Since the actual BOP $P_0^*$ corresponding to embedding {$f_0^*$} is unknown, this method is named ``Shadow Star'', denoted as ``SS''.
By doing so, it is expected that the modified feature embedding of BOP {$f_0^*$} is less out-of-distribution than the original one.
Meanwhile, the resulting start pose of generated action would be closer to the end pose of the previous action compared to naively replacing the original embedding with the most similar one.

\emph{Recognition based ``Action Revision''}
Although Shadow Start can improve the robustness (in terms of action recognition accuracy) of long-term motion generation, it also causes jittering effects due to the modification of the BOP feature. 
Since our method can directly synthesize correct actions in most cases.
therefore, a more practical way is to only use Shadow Start when {~\thename} fails to generate the correct actions.
Specifically, we first generate one action by {\thename}, then an action classifier denoted as $F_c$ is used to classify whether the motion is the desired action. 
If the result is not, we regenerate this action only once with the aforementioned Shadow Start operation. 
We repeat this process during long-term motion generation and call it an ``Action Revision'' scheme, denoted as ``AR''. 
With the above two parts, we summarize the whole {\systemname} in Algorithm ~\ref{algo:hybrid_pip}.
\begin{algorithm}[!h]
\SetAlgoLined
\SetAlFnt{\tiny}
\SetKwInOut{Input}{Input}\SetKwInOut{Output}{Output}
\Input{$\{A_i,T_i,U_i\}_N $,  $\{P_t\}_{t=-x}^{t=-1}$}
\For{$i \gets 1$ \KwTo $N$}{
     $R_z ~ \text{and} ~ D_t \gets P_{-1}$ \;
    Normalized initial pose  $\{\,P^*_t\,\}_{t=-x}^{t=-1} \gets R_z^{-1} \{P_t\}_{t=-x}^{t=-1} - D_t$  \;
    Generate $i$-th action  $\{\,\tilde{P_t}\,\}_{t=1}^{t=T_i} \gets $ {\thename}$(\{P^*\}_{t=-x}^{t=-1},A_i,A_{i+1},T_i,U_i)$\;
    \If{$F_c$ ($ \{\,\tilde{P_t}\,\}_{t=1}^{t=T_i} $) is not correct}
    {
        $f_0^* \gets$ ``Shadow Start'' \;
        Re-generate $i$-th action  with $f_0^*$ \;
    }
    Rotated i-th action $\{\,\hat{P_t}\,\}_{t=1}^{t=T_i} \gets  R_z \tilde{P_t} + D_t $ \;
    Update initial pose sequence  $\{\,P_t\,\}_{t=-x}^{t=-1} \gets 
    \{\, \hat{P_t}\,\}_{t=T_i-x+1}^{t=T_i}$ \;
    }
\caption{\systemname}
\label{algo:hybrid_pip}
\end{algorithm}\vspace{-10pt}

\section{Experiments}\label{sec:experiment}

We conduct experiments for two main tasks: single-action generation and multi-action generation. 
%
%
For the single-action task, we first compare \thename~ with other baselines. 
Besides, we further analyze component effectiveness and controllability of \thename.
%
For the multi-action task, we first compare different generative models under a purely generative pipeline. Both objective and subjective measurements are provided for performance evaluation. %
Then we show how our \systemname~ can further improve the performance of multi-action synthesis.  

\subsection{Single-Action Generation}

\subsubsection{Experiment setup}
\emph{Train/test split.} For every action class, we randomly select $20\%$ to $100$ samples to construct a testing set and use the rest data for training.

\emph{Compared methods.} We consider two state-of-the-art methods, which are carefully adapted and trained on {\dtname}, including \textbf{MANN}~\cite{zhang2018mann}  and  \textbf{Action2motion (A2M)}. Besides, we provide the result of \textbf{ACTOR}~\cite{actor} as a reference since it cannot take the initial poses as input, which makes it unsuitable to generate continuous actions one by one.

\subsubsection{Evaluation metric}
\label{sec:single_methods:eval}
We first consider the commonly used \emph{action-level} metrics to measure the recognition accuracy, realism, and diversity of the synthesized single-action motions~\cite{lee2019dancing,actor,guo2020action2motion}. 
Specifically, a Transformer-based action recognition network $\mathcal{N}$ pre-trained on {\dtname} is used to extract action features. Besides, another pose-level metric is also included to provide a more comprehensive evaluation.  
\begin{itemize}
[leftmargin=1em]
    \item \textbf{Recognition Accuracy (Acc).} We use the pre-trained network $\mathcal{N}$~\cite{guo2020action2motion} to classify the synthesized actions and report the recognition accuracy for each method. 

    \item \textbf{Frechet Inception Distance (FID).} FID~\cite{heusel2017gans} measures the similarity between the distributions of action features (extracted via $\mathcal{N}$) from real-world samples and those from synthetic samples. 
    A \emph{lower} FID suggests a better result when comparing different methods.

    \item \textbf{Diversity (Div).} We measure the \emph{variance} of the features $f$ (extracted via $\mathcal{N}$) of the generated actions across all action categories. 
    We evenly split the generated actions into two subsets $\{f_i\}$ and $\{\tilde{f_i}\}$, and approximate the variance by $\frac{1}{K} \sum_{i=1}^K \big\Vert f_i - \tilde{f_i} \big\Vert_2$ as in~\cite{guo2020action2motion}.
    A \emph{higher} value means the test motions are more diverse. 

    \item \textbf{Multimodality (Multi-mod).} Multimodality measures the diversity \emph{within each action category}. 
    Specifically,  we first compute \emph{variance} of the motion features (via $\mathcal{N}$) of the generated actions sharing the same action tag, then average the variances over different action categories. 
    A \emph{higher} value suggests a larger inter-class diversity.
    \item \textbf{Trajectory Error (\boldsymbol{$\Delta_{Traj}$}).} We measure the trajectory error by calculating the L2 distance between the input trajectory and the generated root position per frame of root translation. 

\end{itemize}

\begin{table}[!ht]
    \centering
    \caption{Comparison on single-action synthesis. }
    \label{tab:res:singl_action}
    \vspace{-10pt}
    \centering
{\def\arraystretch{1}\tabcolsep=0.5em 
\begin{tabular}{lcccccc}
\toprule[1pt]
\multicolumn{1}{c}{Metrics} & FID $\downarrow$ & Acc $\uparrow$ & Div $\uparrow$ & Multi-m $\uparrow$ & {$\Delta_{Traj}$} $\downarrow$
\\ \midrule[1pt]
\textbf{Real Test Data} & \textbf{/}  & \textbf{0.894} & \textbf{9.865} & \textbf{1.956}  & \textbf{/}   \\ 
\midrule[0.5pt]

MANN~\cite{zhang2018mann} & \underline{3.406} & \underline{0.838} & \underline{9.789} & \underline{2.079} &\underline{0.154} \\
A2M~\cite{guo2020action2motion} & 3.152 & 0.826  & 9.738 & 2.048 & /\\

\textbf{Ours } &
{\textbf{1.115}}  & {\textbf{0.874}}  & {\textbf{9.904}} & 2.068  & {\textbf{0.129}}  \\
\midrule[0.5pt]

ACTOR~\cite{actor} & 9.558 & 0.744   & 9.424 & {\textbf{2.356}} & 0.325\\

\bottomrule[1pt]
\end{tabular}
}

\end{table}
\subsubsection{Comparison}
\label{single:res}
\begin{figure}[!h]
    \centering
    \begin{overpic}[width=1\linewidth]{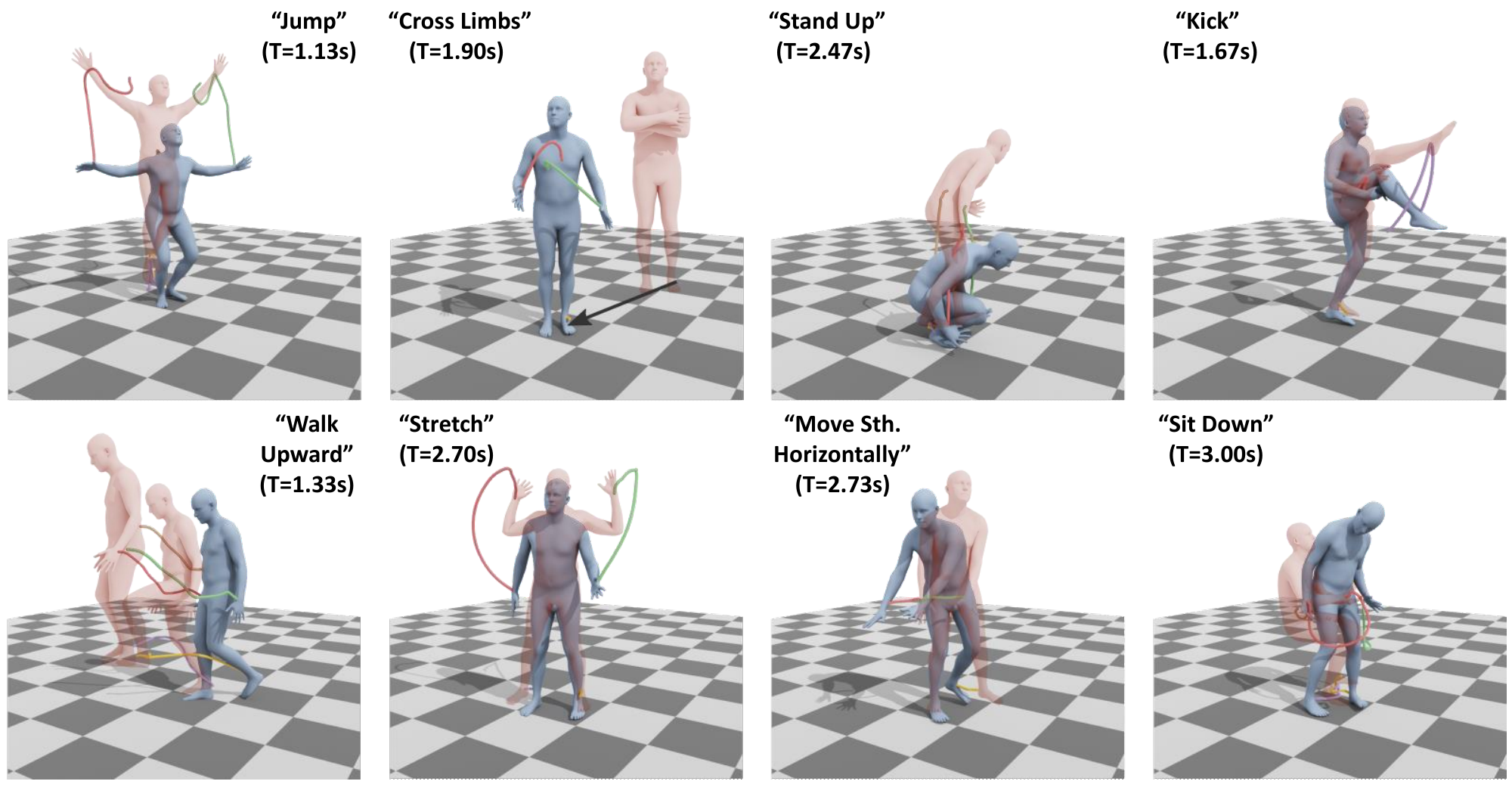}
    
    
    
    

    

    
    \end{overpic}
    \vspace{-25pt}
    \caption{We show eight synthesized actions with different tags using our {\thename}, including the first pose (blue mesh), pose in keyframes (red meshes), and joint trajectories (lines in different colors). \vspace{-10pt}
    }
    \label{fig:res:singleAction:egs}
\end{figure}

In this task, a method is required to generate actions according to the provided action tag.
MANN, A2M, and ours use both action tag and contextual information as input while ACTOR takes only the action tag.
Results of all metrics are presented in Table~\ref{tab:res:singl_action}.
Our method obtains a significantly lower (better) FID score (\boldsymbol{${1.1}$}) among all methods, indicating better motion quality. A visual comparison result is presented in Fig.~\ref{fig:res:singleAction:egs}.
%
For the recognition evaluation, our method achieves the accuracy of \boldsymbol{${0.87}$}, which is higher than the others and approaches the level of real data.
It shows that our method can generate motions accurately corresponding to the given action tag.
As for diversity measurements, three methods considering contextual information achieve similar scores while our method obtains the highest diversity score with \boldsymbol{$9.9$} which is even more diverse than the real data marginally.
MANN outperforms the other two methods on multimodality metric with the value \boldsymbol{$2.08$}, while our method obtains a very closing value of \boldsymbol{$2.07$}. 
In addition to these action-level metrics, {\thename} also shows the best performance in terms of Trajectory Error among all methods, which indicates our method has a better trajectory control ability.  

We could easily find from Table~\ref{tab:res:singl_action} that ACTOR, as the state-of-the-art action-conditioned generative model, is still hard to generate tag-desired and high-quality motions without using any contextual information in our dataset.
Note that, Diversity and multimodality are meaningful only when high-quality and tag-preferred motions are provided. 
Therefore, the highest multimodality score achieved by ACTOR is not necessarily the inter-class diversity that we need.
In summary, our method outperforms other baselines on overall quality and action accuracy with a substantial margin and has a similar performance to other baselines on diversity and multimodality.

\subsubsection{Ablation Study}
We ablate several important components of our method to verify their effectiveness, including Schedule Sampling strategy, Time Unrolling (\emph{TU}), and Tag Injection (\emph{TI}).
All the ablative experiments (see Table~\ref{tab:res:ablation_network}) are conducted under the single-action generation setting.
First of all, {\thename} trained without the scheduled sampling suffers from larger Joint Position Error inevitably while our proposed training strategy greatly narrows the train-test discrepancy. 

Besides, we find that both Time Unrolling and Tag Injection contribute to improving the quality of generated motions in terms of all metrics except multimodality. 
At the same time, the former has a larger impact than the latter on FID, accuracy, and Joint Position Error, which shows that unrolling latent code $Z$ to time-varying sequential features plays a key role in our generative model and can significantly improve naturalness and accuracy of the generated actions.

\begin{table}[!t]
    \centering
    \caption{Ablation study of key components of {\thename}.}
    \label{tab:res:ablation_network}\vspace{-10pt}
    \centering
{\def\arraystretch{1}\tabcolsep=0.2em 
\begin{tabular}{lcccccc}
\toprule[1pt]
\multicolumn{1}{c}{Metrics} & FID $\downarrow$ & Acc $\uparrow$ & Div $\uparrow$ & Multi-mod $\uparrow$ & {$\Delta_{JPE}$} $\downarrow$
\\ \midrule[1pt]

\textbf{Ours ({\thename})} &
{\textbf{1.115}}  & {\textbf{0.874}}   & {\textbf{9.904}} & 
2.068 & 
{\textbf{4.58}} \\
\midrule[0.5pt]

\textit{w/o Schedule Sampling} &
\underline{1.274}& \underline{0.871}  & 9.846  & 1.985 & \underline{6.99} \\
\midrule[0.5pt]

\textit{w/o TU and w/o TI} 
& 11.294  & 0.638  & 9.706 
& \underline{2.404} &9.32\\

\textit{w/o TU} 
& 11.522  & 0.676 & \underline{9.867} 
& {\textbf{2.442}}&9.42\\
\textit{w/o TI} 
& 5.040 & 0.819 & 9.458 
& 2.284 &7.84\\

\bottomrule[1pt]
\end{tabular}
}




\end{table}

\begin{figure}[!h]
    \centering
    \includegraphics[width=0.5\textwidth]{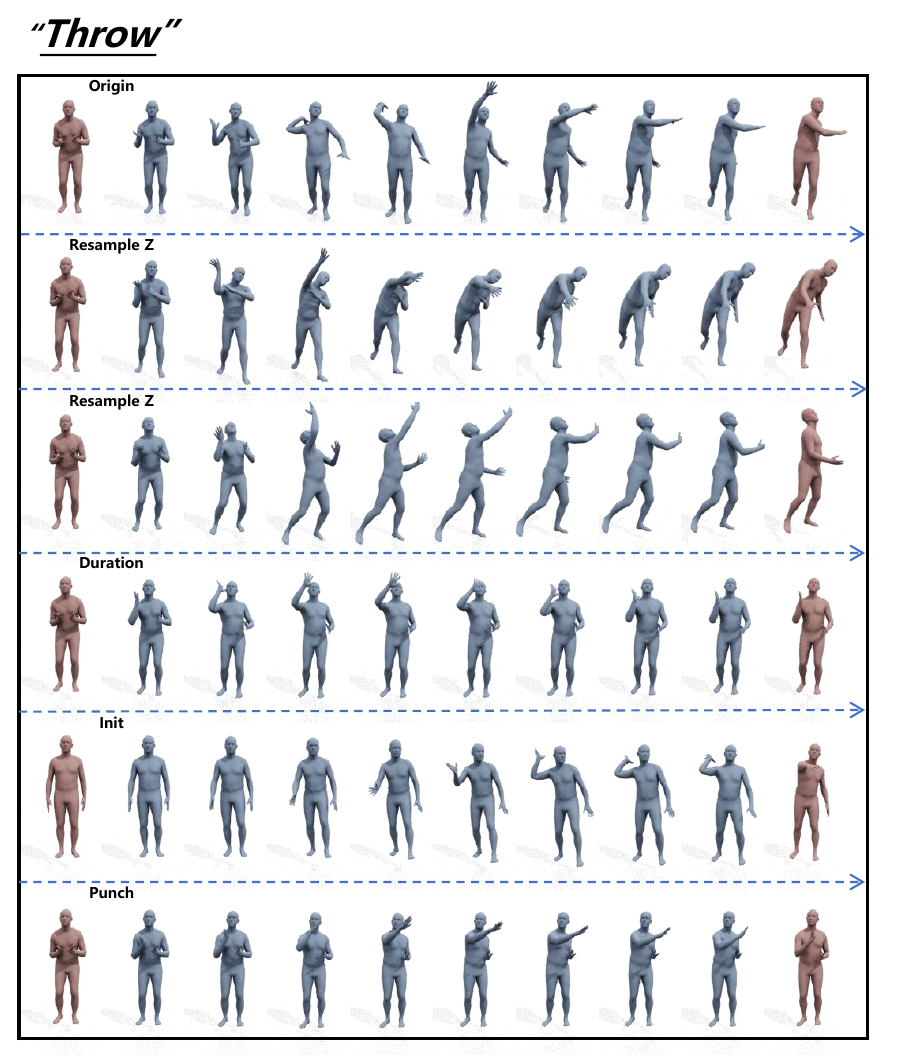}\vspace{-15pt}
    \caption{Effectiveness of different control signals.  The input of the 1st row is from the real action of ``throw''. Compared to the 1st row: (2),(3) re-samples condition latent code $Z$; (4) changes duration (5) changes initial pose (6) changes action tag to ``Punch''. It can be clearly seen that actions are different from each other.}\vspace{-10pt}
    \label{fig:singal_change1}
\end{figure}

\subsubsection{Qualitative results of model controllability}
As a conditional generative model, {\thename} can produce diverse actions corresponding to different inputs.
In this part, we present a visual example in Fig.~\ref{fig:singal_change1} to demonstrate the controllability of {\thename}.
The first row of this example is base-generated actions, of which all input conditions are from original real data including action tag (``throw''), duration, initial motion, and trajectory. The latent code $Z$ is directly set to the predicted $\mu$.
The second and the third rows show actions generated with different re-sampling latent code $Z$.
It can be seen that these three actions (1st row to 3rd row) are significantly different, which demonstrates that, even with the same input, our model still can generate diverse motions belonging to the same action category. 
For each row from the fourth to the sixth row, we show generated actions with only one different input control signal compared to the first row. 
 It can be easily found that all these actions are different as expected. Especially, for the last row, we change action tags (change ``throw'' to ``punch'' for Fig~\ref{fig:singal_change1}). 
Even with the same initial motion,  {\thename} can generate different action types according to the input action tags.

\subsection{Multi-Action Generation}

%
In this part, we are focusing on evaluating multi-action generation performance. 
%
%
For conducting more comprehensive experiments, we firstly construct two multi-action testing sets from the BABEL dataset by considering different connection frequencies between actions. 
Based on testing sets, we compare our {\thename} with two adapted state-of-the-art action generation methods under the pure generative pipeline. Both subjective and objective evaluations are provided to discuss the performance of multi-action synthesis.
Lastly, we demonstrate how the hybrid system, {\systemname} can further improve performance compared to the pure generative pipeline. 

%

\subsubsection{Construction of multi-action evaluation sets}
For this task, we need to prepare a sequence of action tags and initial motion as inputs.
Directly selecting an animation from BABEL may include extra actions that are not labeled in our dataset.
Instead, we create multi-action animations by connecting actions from the single-action task test set according to the transition matrix where the number of connections between different action categories in the training set is calculated in advance (see Fig. 1 in Supp.). 
Please note these multi-action animations are not continual motions in the original data, and we only use their conditional information as inputs, including action tags, duration, and trajectories (for ``root motion'' only).

In order to analyze the influence of the number of transitions between action categories that existed in training data, we set up two kinds of multi-action generation evaluation sets, namely ``overall'' and ``sufficient'' testing. The ``overall'' set is obtained by randomly connected action tags which are at least connected once in the training set.  The ``sufficient'' set only includes action tag sequences that have frequently connected in the training set. Every set has \boldsymbol{$764$} multi-action samples that contain \boldsymbol{$20$} actions. (Detailed construction steps of these two sets are provided in Supplementary.).

\subsubsection{Evaluation metrics}
It has not been well established that the criteria to evaluate the multi-action motion generation task.
%
In this paper, we suggest evaluating multi-action animation from three aspects:
\begin{itemize}
[leftmargin=*]
    \item \textbf{Per-action Quality}: we measure action recognition accuracy and FID (See Sec.~\ref{sec:single_methods:eval}) for action clips which are split from generated results. These two metrics are presented to jointly demonstrate the quality of synthetic actions with the same sequential index. 
    \item \textbf{Transition Smoothness}: we evaluate transition smoothness between two actions based on pose and velocity gaps.
    For \emph{pose gap} ($\Delta_{pose}$, in $cm $), we use the L2 distance of per-joint position between the end pose of the previous action and the start one in the current action. 
    A lower value usually indicates a smoother transition.
    For \emph{velocity gap} ($\Delta_{vel}$, in $cm/frame$), we also measure the difference of per-joint velocity between the end of the previous action and the start of the current action, which evaluates the dynamic smoothness of transitions.
    \item \textbf{Qualification Ratio}: We define action-level and motion-level qualification ratio (QR) as follows.
    For action-level QR, we calculate the percentage of accurately classified actions among all independent actions in a generated motion sequence. 
    Given an example, a motion including $3$ accurate actions among total $5$ actions, then its action-level QR is $60\%$.
    We report the average of all testing data as action-level QR for every method.
    For motion-level QR, we define the animation as a qualified one only if all actions are recognized correctly.
    Then the Motion QR can be calculated simply as:
    \begin{equation}
        \text{Motion QR} = \frac{\text{number of qualified animations}}{\text{number of all animations}}.
    \end{equation}
    Motion-level QR represents the quality of the full motion sequence, which is a much more strict measurement than action-level QR.
\end{itemize}

\begin{figure}[!h]
    \centering
    \includegraphics[width=0.5\textwidth,height=0.9\linewidth]{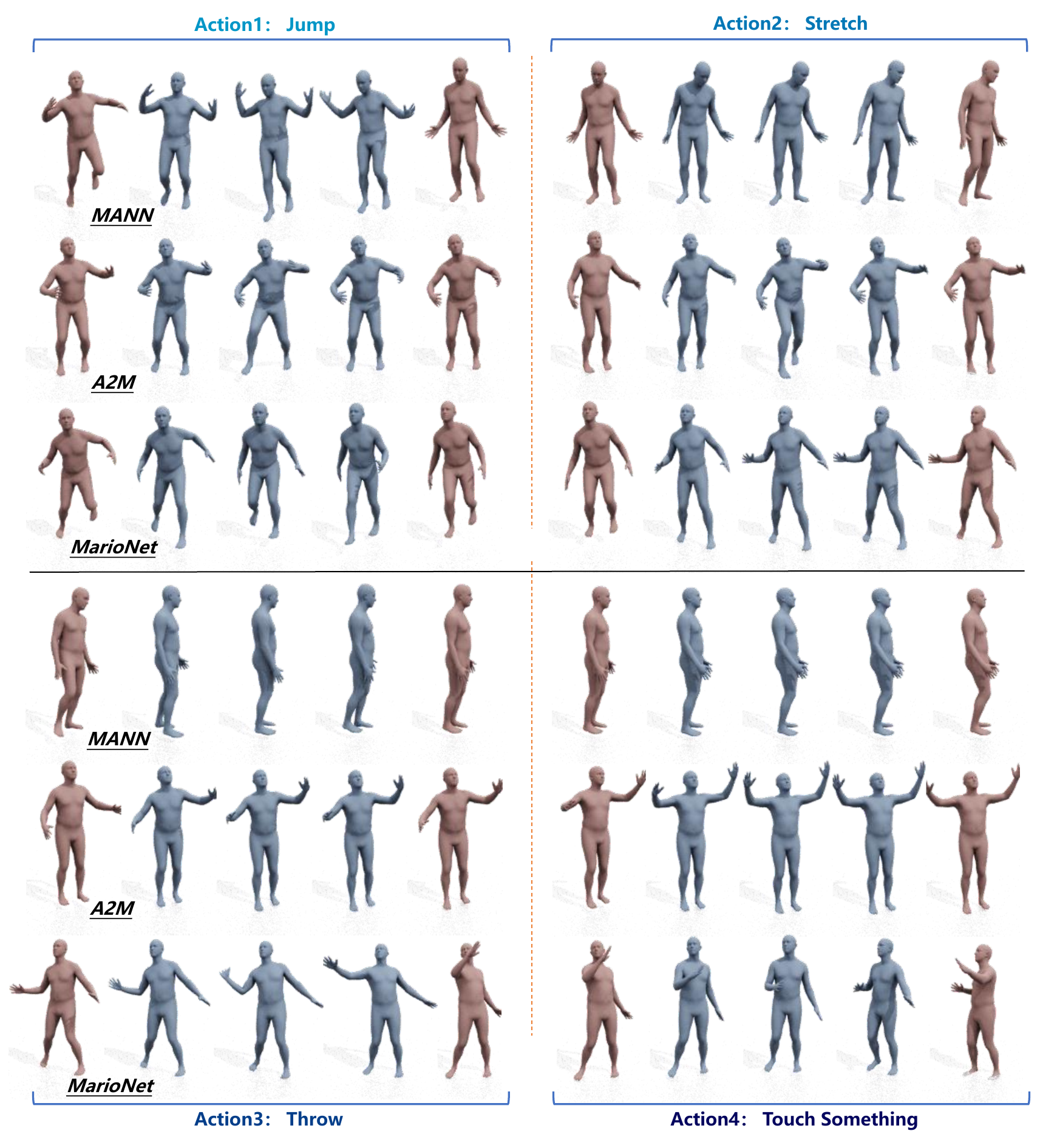}\vspace{-14pt}
    \caption{Action visualization of multi-action generation by different methods.  Severe abrupt pose changes between actions could be easily observed in A2M, while MANN does not generate correct desired actions. Our {\thename} successfully generates expected actions with a smooth transition.}
    \label{fig:multi_compare}
\end{figure}
\begin{figure*}[!t]
\begin{tikzpicture}[scale=1.0]
\pgfplotsset{
compat=1.11,
legend image code/.code={
\draw[mark repeat=2,mark phase=2]
plot coordinates {
(0cm,0cm)
(0.22cm,0.0cm)        
(0.1cm,0cm)         
};%
}
}
\pgfmathsetlengthmacro\MajorTickLength{
      \pgfkeysvalueof{/pgfplots/major tick length} * 0.5
    }
\definecolor{dimgray85}{RGB}{85,85,85}
\definecolor{gainsboro229}{RGB}{229,229,229}
\definecolor{gray}{RGB}{28,28,28}
\definecolor{lightgray}{RGB}{120,120,120}
\definecolor{myred}{HTML}{ff595e}
\definecolor{mygreen}{HTML}{8ac926}
\definecolor{myblue}{HTML}{4895ef}

\begin{axis}[
width=0.28\linewidth,
axis background/.style={fill=white},
axis line style={gray},
legend cell align={left},
legend columns=3,
legend style={
  fill opacity=0.8,
  draw opacity=1,
  text opacity=1,
  at={(0.51,-0.015)},
  anchor=south,
  draw=none,
  fill=none,
  font=\tiny
},
tick align=inside,
tick pos=left,
major tick length=\MajorTickLength,
title style={font=\bfseries, at={(0.5,0.95)}},
title={(a)},
x grid style={gray!10},
xlabel style={at={(axis description cs:0.5,-0.1)},font=\footnotesize\itshape},
xlabel=\textcolor{gray}{Index of action},
xmajorgrids,
xmin=1, xmax=20,
xtick style={color=dimgray85},
y grid style={gray!10},
ylabel style={at={(axis description cs:-0.12,.5)},font=\footnotesize\itshape},
ylabel=\textcolor{gray}{Accuracy},
ymajorgrids,
ymin=0.6, ymax=1,
ytick style={color=dimgray85},
tick label style={font=\footnotesize}
]
\addplot [  line width=1.0pt, myred]
table {%
1 0.945
2 0.853
3 0.825
4 0.819
5 0.797
6 0.802
7 0.822
8 0.779
9 0.777
10 0.781
11 0.797
12 0.749
13 0.763
14 0.747
15 0.777
16 0.781
17 0.76
18 0.779
19 0.774
20 0.747
};
\addlegendentry{\underline{MarioNet-O}}

\addplot [  line width=1.0pt, mygreen]
table {%
1 0.889
2 0.801
3 0.747
4 0.767
5 0.733
6 0.729
7 0.743
8 0.728
9 0.728
10 0.703
11 0.707
12 0.699
13 0.699
14 0.721
15 0.717
16 0.694
17 0.705
18 0.723
19 0.7
20 0.691
};
\addlegendentry{MANN-O}

\addplot [  line width=1.0pt, myblue]
table {%
1 0.891
2 0.814
3 0.805
4 0.788
5 0.781
6 0.764
7 0.775
8 0.755
9 0.791
10 0.758
11 0.763
12 0.77
13 0.771
14 0.779
15 0.781
16 0.789
17 0.766
18 0.751
19 0.76
20 0.772
};
\addlegendentry{A2M-O}

\addplot [  line width=1.0pt, myred, dashed]
table {%
1 0.945
2 0.882
3 0.863
4 0.864
5 0.868
6 0.856
7 0.872
8 0.843
9 0.861
10 0.864
11 0.861
12 0.852
13 0.852
14 0.848
15 0.865
16 0.872
17 0.869
18 0.881
19 0.868
20 0.843
};
\addlegendentry{\underline{MarioNet-S}}
\addplot [  line width=1.0pt, mygreen, dashed]
table {%
1 0.889
2 0.806
3 0.774
4 0.758
5 0.75
6 0.758
7 0.759
8 0.737
9 0.77
10 0.772
11 0.755
12 0.747
13 0.745
14 0.745
15 0.726
16 0.711
17 0.707
18 0.729
19 0.717
20 0.709
};
\addlegendentry{MANN-S}

\addplot [  line width=1.0pt, myblue, dashed]
table {%
1 0.891
2 0.848
3 0.859
4 0.832
5 0.851
6 0.819
7 0.855
8 0.831
9 0.839
10 0.843
11 0.827
12 0.813
13 0.829
14 0.814
15 0.825
16 0.823
17 0.804
18 0.81
19 0.819
20 0.83
};
\addlegendentry{A2M-S}
\addplot [  line width=0.5pt, gray, dashed, forget plot]
table {%
2 0.6
2 1
};
\end{axis}

\end{tikzpicture} \hspace{-12pt}
\begin{tikzpicture}[scale=1.0]
\pgfmathsetlengthmacro\MajorTickLength{
      \pgfkeysvalueof{/pgfplots/major tick length} * 0.5
    }
\definecolor{dimgray85}{RGB}{85,85,85}
\definecolor{gainsboro229}{RGB}{229,229,229}
\definecolor{gray}{RGB}{28,28,28}
\definecolor{lightgray}{RGB}{120,120,120}
\definecolor{myred}{HTML}{ff595e}
\definecolor{mygreen}{HTML}{8ac926}
\definecolor{myblue}{HTML}{4895ef}

\begin{axis}[
width=0.28\linewidth,
axis background/.style={white},
axis line style={gray},
legend cell align={left},
legend columns=3,
legend style={
  fill opacity=0.8,
  draw opacity=1,
  text opacity=1,
  at={(0.5,0.09)},
  anchor=south,
  draw=lightgray,
  fill=gainsboro229
},
tick align=inside,
tick pos=left,
major tick length=\MajorTickLength,
title style={font=\bfseries, at={(0.5,0.95)}},
title={(b)},
x grid style={gray!10},
xlabel style={at={(axis description cs:0.5,-0.1)},font=\footnotesize\itshape},
xlabel=\textcolor{gray}{Index of action},
xmajorgrids,
xmin=1, xmax=20,
xtick style={color=dimgray85},
y grid style={gray!10},
ylabel style={at={(axis description cs:-0.09,.5)},font=\footnotesize\itshape},
ylabel=\textcolor{gray}{FID},
ymajorgrids,
ymin=0, ymax=22,
ytick style={color=dimgray85},
tick label style={font=\footnotesize}
]
\addplot [ line width=1.0pt, myred]
table {%
1 1.178
2 6.926
3 6.789
4 6.353
5 7.159
6 7.109
7 7.654
8 8.509
9 8.196
10 8.54
11 8.279
12 8.566
13 8.229
14 7.729
15 8.564
16 8.754
17 8.074
18 7.713
19 7.63
20 8.456
};
\addplot [ line width=1.0pt, myred, dashed]
table {%
1 1.178
2 3.191
3 3.354
4 4.232
5 4.555
6 4.592
7 4.287
8 4.545
9 4.602
10 4.667
11 4.53
12 4.541
13 5.455
14 4.843
15 5.731
16 5.163
17 4.865
18 4.614
19 5.139
20 4.952
};
\addplot [ line width=1.0pt, mygreen]
table {%
1 4.317
2 10.19
3 11.289
4 11.023
5 11.625
6 12.266
7 12.836
8 13.062
9 12.645
10 14.729
11 13.076
12 13.401
13 13.371
14 12.498
15 13.41
16 13.118
17 12.594
18 14.355
19 13.935
20 13.484
};
\addplot [ line width=1.0pt, mygreen, dashed]
table {%
1 4.317
2 7.519
3 9.154
4 11.267
5 12.549
6 14.027
7 14.426
8 15.834
9 16.421
10 15.454
11 15.282
12 16.232
13 16.568
14 15.965
15 17.429
16 16.28
17 16.5
18 16.388
19 16.191
20 16.646
};
\addplot [ line width=1.0pt, myblue]
table {%
1 3.846
2 11.21
3 12.1
4 11.749
5 12.695
6 13.776
7 14.105
8 14.194
9 14.361
10 16.07
11 16.425
12 16.391
13 16.454
14 16.846
15 17.585
16 19.598
17 18.233
18 18.873
19 19.351
20 20.116
};
\addplot [ line width=1.0pt, myblue, dashed]
table {%
1 3.846
2 6.264
3 6.575
4 7.357
5 6.439
6 7.04
7 6.834
8 7.573
9 8.007
10 7.468
11 8.1
12 8.843
13 9.094
14 8.913
15 9.051
16 8.326
17 9.023
18 8.65
19 8.481
20 8.681
};
\addplot [ line width=0.5pt, gray, dashed, forget plot]
table {%
2 0
2 22
};
\end{axis}

\end{tikzpicture} \hspace{-12pt}
\begin{tikzpicture}[scale=1.0]
\pgfmathsetlengthmacro\MajorTickLength{
      \pgfkeysvalueof{/pgfplots/major tick length} * 0.5
    }
\definecolor{dimgray85}{RGB}{85,85,85}
\definecolor{gainsboro229}{RGB}{229,229,229}
\definecolor{gray}{RGB}{28,28,28}
\definecolor{lightgray}{RGB}{120,120,120}
\definecolor{myred}{HTML}{ff595e}
\definecolor{mygreen}{HTML}{8ac926}
\definecolor{myblue}{HTML}{4895ef}

\begin{axis}[
width=0.28\linewidth,
axis background/.style={fill=white},
axis line style={gray},
legend cell align={left},
legend columns=3,
legend style={
  fill opacity=0.8,
  draw opacity=1,
  text opacity=1,
  at={(0.5,0.04)},
  anchor=south,
  draw=none,
  fill=none,
  line width=1.0pt
},
tick align=inside,
tick pos=left,
major tick length=\MajorTickLength,
title style={font=\bfseries, at={(0.5,0.95)}},
title={(c)},
x grid style={gray!10},
xlabel style={at={(axis description cs:0.5,-0.1)},font=\footnotesize\itshape},
xlabel=\textcolor{gray}{Number of actions},
xmajorgrids,
xmin=1, xmax=20,
xtick style={color=dimgray85},
y grid style={gray!10},
ylabel style={at={(axis description cs:-0.12,.5)},font=\footnotesize\itshape},
ylabel=\textcolor{gray}{Action QR},
ymajorgrids,
ymin=0.6, ymax=1,
ytick style={color=dimgray85},
tick label style={font=\footnotesize}
]
\addplot [line width=1.0pt, myred]
table {%
1 0.945
2 0.899
3 0.874333333333333
4 0.8605
5 0.8478
6 0.840166666666667
7 0.837571428571429
8 0.83025
9 0.824333333333333
10 0.82
11 0.817909090909091
12 0.812166666666667
13 0.808384615384616
14 0.804
15 0.8022
16 0.800875
17 0.798470588235294
18 0.797388888888889
19 0.796157894736842
20 0.7937
};
\addplot [line width=1.0pt, mygreen]
table {%
1 0.889
2 0.845
3 0.812333333333333
4 0.801
5 0.7874
6 0.777666666666667
7 0.772714285714286
8 0.767125
9 0.762777777777778
10 0.7568
11 0.752272727272727
12 0.747833333333333
13 0.744076923076923
14 0.742428571428571
15 0.740733333333333
16 0.7378125
17 0.735882352941176
18 0.735166666666667
19 0.733315789473684
20 0.7312
};
\addplot [line width=1.0pt, myblue]
table {%
1 0.891
2 0.8525
3 0.836666666666667
4 0.8245
5 0.8158
6 0.807166666666667
7 0.802571428571429
8 0.796625
9 0.796
10 0.7922
11 0.789545454545455
12 0.787916666666667
13 0.786615384615385
14 0.786071428571429
15 0.785733333333333
16 0.7859375
17 0.784764705882353
18 0.782888888888889
19 0.781684210526316
20 0.7812
};
\addplot [line width=1.0pt, myred, dashed]
table {%
1 0.945
2 0.9135
3 0.896666666666667
4 0.8885
5 0.8844
6 0.879666666666667
7 0.878571428571428
8 0.874125
9 0.872666666666667
10 0.8718
11 0.870818181818182
12 0.86925
13 0.867923076923077
14 0.8665
15 0.8664
16 0.86675
17 0.866882352941177
18 0.867666666666667
19 0.867684210526316
20 0.86645
};
\addplot [line width=1.0pt, mygreen, dashed]
table {%
1 0.889
2 0.8475
3 0.823
4 0.80675
5 0.7954
6 0.789166666666667
7 0.784857142857143
8 0.778875
9 0.777888888888889
10 0.7773
11 0.775272727272727
12 0.772916666666667
13 0.770769230769231
14 0.768928571428572
15 0.766066666666667
16 0.762625
17 0.759352941176471
18 0.757666666666667
19 0.755526315789474
20 0.7532
};
\addplot [line width=1.0pt, myblue, dashed]
table {%
1 0.891
2 0.8695
3 0.866
4 0.8575
5 0.8562
6 0.85
7 0.850714285714286
8 0.84825
9 0.847222222222222
10 0.8468
11 0.845
12 0.842333333333333
13 0.841307692307692
14 0.839357142857143
15 0.8384
16 0.8374375
17 0.835470588235294
18 0.834055555555556
19 0.833263157894737
20 0.8331
};
\addplot [line width=0.5pt, gray, dashed, forget plot]
table {%
2 0.6
2 1
};
\end{axis}

\end{tikzpicture} \hspace{-12pt}
\begin{tikzpicture}[scale=1.0]
\pgfmathsetlengthmacro\MajorTickLength{
      \pgfkeysvalueof{/pgfplots/major tick length} * 0.5
    }
\definecolor{dimgray85}{RGB}{85,85,85}
\definecolor{gainsboro229}{RGB}{229,229,229}
\definecolor{gray}{RGB}{28,28,28}
\definecolor{lightgray}{RGB}{120,120,120}
\definecolor{myred}{HTML}{ff595e}
\definecolor{mygreen}{HTML}{8ac926}
\definecolor{myblue}{HTML}{4895ef}

\begin{axis}[
width=0.28\linewidth,
axis background/.style={fill=white},
axis line style={gray},
legend cell align={left},
legend columns=3,
legend style={
  fill opacity=0.8,
  draw opacity=1,
  text opacity=1,
  at={(0.5,0.09)},
  anchor=south,
  draw=none,
  fill=none
},
tick align=inside,
tick pos=left,
major tick length=\MajorTickLength,
title style={font=\bfseries, at={(0.5,0.95)}},
title={(d)},
xlabel style={at={(axis description cs:0.5,-0.1)},font=\footnotesize\itshape},
xlabel=\textcolor{gray}{Number of actions},
xmajorgrids,
xmin=1, xmax=20,
xtick style={color=dimgray85},
y grid style={gray!10},
ylabel style={at={(axis description cs:-0.12,.5)},font=\footnotesize\itshape},
ylabel=\textcolor{gray}{Motion QR},
ymajorgrids,
ymin=0, ymax=1,
ytick style={color=dimgray85},
tick label style={font=\footnotesize}
]
\addplot [line width=1.0pt, myred]
table {%
1 0.945
2 0.827
3 0.724
4 0.661
5 0.572
6 0.525
7 0.475
8 0.436
9 0.391
10 0.365
11 0.343
12 0.297
13 0.266
14 0.236
15 0.219
16 0.191
17 0.171
18 0.162
19 0.145
20 0.127
};
\addplot [line width=1.0pt, mygreen]
table {%
1 0.889
2 0.754
3 0.644
4 0.575
5 0.503
6 0.432
7 0.382
8 0.339
9 0.292
10 0.255
11 0.223
12 0.204
13 0.185
14 0.162
15 0.147
16 0.136
17 0.13
18 0.116
19 0.101
20 0.084
};
\addplot [line width=1.0pt, myblue]
table {%
1 0.891
2 0.763
3 0.666
4 0.59
5 0.52
6 0.473
7 0.437
8 0.395
9 0.368
10 0.353
11 0.325
12 0.312
13 0.293
14 0.284
15 0.277
16 0.272
17 0.266
18 0.255
19 0.247
20 0.243
};
\addplot [line width=1.0pt, myred, dashed]
table {%
1 0.945
2 0.853
3 0.785
4 0.733
5 0.7
6 0.665
7 0.626
8 0.598
9 0.576
10 0.55
11 0.522
12 0.497
13 0.479
14 0.458
15 0.448
16 0.435
17 0.42
18 0.412
19 0.395
20 0.376
};
\addplot [line width=1.0pt, mygreen, dashed]
table {%
1 0.889
2 0.766
3 0.681
4 0.603
5 0.53
6 0.49
7 0.454
8 0.419
9 0.389
10 0.376
11 0.338
12 0.313
13 0.285
14 0.26
15 0.245
16 0.215
17 0.198
18 0.187
19 0.169
20 0.156
};
\addplot [line width=1.0pt, myblue, dashed]
table {%
1 0.891
2 0.792
3 0.738
4 0.692
5 0.664
6 0.622
7 0.598
8 0.573
9 0.551
10 0.534
11 0.509
12 0.479
13 0.458
14 0.44
15 0.421
16 0.407
17 0.386
18 0.38
19 0.365
20 0.349
};
\addplot [line width=0.5pt, gray, dashed, forget plot]
table {%
2 0
2 1
};
\end{axis}

\end{tikzpicture} \vspace{-28pt}
    \caption{Quantitative results for overall (solid lines and denoted with ``-O'' ) and sufficient (dashed lines and denoted with ``-S'')  testing sets.   }\vspace{-10pt}
    \label{fig:res:mag_acc_fid}
\end{figure*}

\subsubsection{Comparison}

We first compare our {\thename} with the other two methods in a purely generative pipeline.
Then we demonstrate how the proposed ``Shadow Start'' and ``Action Revision'' schemes can further improve the performance of multi-action generation.
For the purely generative pipeline, the whole motions are only synthesized by generative models sequentially, detailed algorithm can be found in Supplementary Algorithm.1.
Generative methods are evaluated with both overall and sufficient testing.
Results are presented in Table~\ref{tab:res:multi_action_all} and Fig.~\ref{fig:res:mag_acc_fid}. Besides, %
we present a qualitative result of multi-action generation by the above three methods in Fig.~\ref{fig:multi_compare}.
Note that the \emph{averaged} scores in the table are calculated from the \emph{second} action to the last one since the initial motion of the first action is from \dtname{}, which is different from later action generation that uses previously generated poses as input.
\begin{table}[!ht]
    \centering
    \caption{Comparison on multi-action synthesis for pure generative methods.}
    \label{tab:res:multi_action_all}\vspace{-10pt}
    \centering
{\def\arraystretch{1}\tabcolsep=0.2em 
\begin{tabular}{@{}lcccccccc@{}}
\toprule[1pt]
 & \multicolumn{4}{c}{Overall}         & \multicolumn{4}{c}{Sufficient}             \\ \cmidrule(lr){2-5} \cmidrule(lr){6-9}
 Method & Acc$\uparrow$    & FID$\downarrow$  & $\Delta_{pose}\downarrow$ & $\Delta_{vel}\downarrow$ & Acc$\uparrow$  & FID$\downarrow$    & $\Delta_{pose}\downarrow$ & $\Delta_{vel}\downarrow$ \\ 
 \midrule[1pt]

%
MANN & 0.723 & \underline{12.785} & \textbf{1.47}    & \underline{0.42}    & 0.746 & 14.744 & \textbf{1.56}    & \underline{0.37}    \\

A2M  & \underline{0.775} & 15.796 & 3.41    & 1.92    & \underline{0.830} & \underline{7.933}  & 4.12    & 1.73   \\ 

\textbf{{\thename}} & \textbf{0.786} & \textbf{7.854}  & \underline{2.03}    & \textbf{0.40}   & \textbf{0.862} & \textbf{4.624}  & \underline{2.53}    & \textbf{0.35}    \\





\bottomrule[1pt]
\end{tabular}
}
\end{table}
First, we compare per-action quality and qualification ratio metrics among different methods.
Under the ``overall testing'' setting, our method obtains the highest Recognition Accuracy, Action QR, Motion QR, and lowest FID (i.e. most realistic motions).
Fig.~\ref{fig:res:mag_acc_fid} shows how these metrics change as more actions are sequentially generated by different methods. 
As expected, the accuracy of all methods decreases, since the ending poses in the previously generated action may have not been seen in the training set.
However, our method consistently performs best on all metrics within $8$ actions. 
After $8$ actions, our method is significantly better in terms of FID and generates multiple actions with smooth transitions while performing similarly with A2M on other metrics, indicating that the realism of our generated actions remains high quality.

Next, we compare the Transition Smoothness by two pose-level metrics among different methods (see Table~\ref{tab:res:multi_action_all}).
For both settings, our method has a slightly larger pose gap than MANN but obtains a smoother dynamical transition with a lower velocity gap.
A2M reports the worst result in terms of the two Transition Smooth metrics over both settings, which is verified in Fig.~\ref{fig:multi_compare} where A2M easily produces abrupt pose changes.

Finally, we can see choosing the next action restricted on the transition connection matrix ( Fig. 1 in Supp.) rankings can improve the performance by \boldsymbol{$8\%$} on both Accuracy and Action QR, reducing/increasing the FID/Motion QR by nearly half(``sufficient'' vs ``overall''). This phenomenon supports that the quality of multi-action generation can be further improved if there are more combinations available between every two action categories during training. More visual results can be found in the supplementary material.

\subsubsection{Effectiveness of Next tag}
\label{sec:eff_NextTag}
To better understand how contextual information of input influences our model, we conduct an ablation study on the next tag information and present results in Tab~\ref{tab:res:noNt}. \textbf{W/O} next tag is denoted that this model is trained with all next tag inputs set to zero. \textbf{W/} next tag is the original method.  It can be observed that, for the single-action generation task, both models have a close performance of accuracy.
However, for the multi-action generation task, the next tag information can help our model achieve higher accuracy. 
Besides, we also notice that the improvement is more obvious for the sufficient test dataset, which means the effectiveness of the next tag information is influenced by the number of connected tag pairs.
The more connected-tag pairs in the training set,  the more the performance is improved by the next tag information.

\begin{table}[!h]
    \centering
    \caption{effectiveness of Next tag. }
    \label{tab:res:noNt}\vspace{-10pt}
    \begin{tabular}{llll}
\hline
\multirow{2}{*}{} & \multicolumn{1}{c}{\multirow{2}{*}{Single-action}} & \multicolumn{2}{c}{Multi-action} \\ 
                  & \multicolumn{1}{c}{}                               & Overall       & Sufficient       \\ \hline
W/              & 0.874                                              & 0.786         & 0.862            \\
W/O             & 0.875                                              & 0.770         & 0.823            \\ \hline
\end{tabular}
\end{table}
%
%

%
\subsubsection{Subjective evaluation}
\label{sec:res_userStudy}

\begin{figure}[!h]
    \pgfplotsset{compat=1.11,
        /pgfplots/my ybar legend/.style={
        /pgfplots/legend image code/.code={%
        \draw[##1,/tikz/.cd,bar width=3pt,yshift=-0.2em,bar shift=0pt]
                plot coordinates {(0cm,0.8em)};},
},
}
\begin{tikzpicture}[scale=0.57]

\definecolor{dimgray85}{RGB}{85,85,85}
\definecolor{gainsboro229}{RGB}{229,229,229}
\definecolor{goldenrod1911910}{RGB}{191,191,0}
\definecolor{lightgray204}{RGB}{204,204,204}
\definecolor{gray}{RGB}{28,28,28}
\definecolor{lightgray}{RGB}{120,120,120}
\definecolor{myred}{HTML}{ff595e}
\definecolor{mygreen}{HTML}{8ac926}
\definecolor{myblue}{HTML}{4895ef}
\definecolor{myyellow}{HTML}{ffbe0b}

\begin{axis}[
width=0.9\linewidth,
height=0.55\linewidth,
axis background/.style={white},
axis line style={gray},
legend cell align={left},
legend columns=4,
legend style={fill opacity=0.8, draw opacity=1, text opacity=1, draw=none, at={(0.5,0.82)},anchor=south, fill=none,font=\footnotesize},
tick align=inside,
tick pos=left,
title style={font=\bfseries, at={(0.5,0.95)}},
title={w/ Real data},
x grid style={white},
xmajorgrids,
xmin=-0.29, xmax=3.89,
xtick style={color=dimgray85},
xtick={0.2,1.2,2.2,3.4},
xticklabel style={font=\bfseries\itshape},
xticklabels={A-1 Acc,A-2 Acc,Smoothness,Naturalism},
y grid style={white},
ymajorgrids,
ymin=0, ymax=1.1,
ytick={0,0.2,0.4,0.6,0.8,1},
ylabel style={font=\bfseries\itshape},
ylabel=\textcolor{gray}{Score},
ytick style={color=dimgray85}
]
\draw[draw=none,fill=myyellow,very thin] (axis cs:-0.1,0) rectangle (axis cs:0.1,0.917857143);
\addlegendimage{ybar,my ybar legend,draw=none,fill=myyellow,very thin}

\addlegendentry{\bfseries\itshape Real data}

\draw[draw=none,fill=myyellow,very thin] (axis cs:0.9,0) rectangle (axis cs:1.1,0.864285714);
\draw[draw=none,fill=myyellow,very thin] (axis cs:1.9,0) rectangle (axis cs:2.1,0.922857143);
\draw[draw=none,fill=myyellow,very thin] (axis cs:2.9,0) rectangle (axis cs:3.1,0.937857143);
\draw[draw=none,fill=myred,very thin] (axis cs:0.1,0) rectangle (axis cs:0.3,0.792857143);
\addlegendimage{ybar,my ybar legend,draw=none,fill=myred,very thin}
\addlegendentry{\bfseries\itshape\underline{MarioNet}}

\draw[draw=none,fill=myred,very thin] (axis cs:1.1,0) rectangle (axis cs:1.3,0.646428571);
\draw[draw=none,fill=myred,very thin] (axis cs:2.1,0) rectangle (axis cs:2.3,0.725714286);
\draw[draw=none,fill=myred,very thin] (axis cs:3.1,0) rectangle (axis cs:3.3,0.692142857);
\draw[draw=none,fill=mygreen,very thin] (axis cs:0.3,0) rectangle (axis cs:0.5,0.378571429);
\addlegendimage{ybar,my ybar legend,draw=none,fill=mygreen,very thin}
\addlegendentry{\bfseries\itshape A2M}

\draw[draw=none,fill=mygreen,very thin] (axis cs:1.3,0) rectangle (axis cs:1.5,0.253571429);
\draw[draw=none,fill=mygreen,very thin] (axis cs:2.3,0) rectangle (axis cs:2.5,0.465714286);
\draw[draw=none,fill=mygreen,very thin] (axis cs:3.3,0) rectangle (axis cs:3.5,0.420714286);
\draw[draw=none,fill=myblue,very thin] (axis cs:0.5,0) rectangle (axis cs:0.7,0.271428571);
\addlegendimage{ybar,my ybar legend,draw=none,fill=myblue,very thin}
\addlegendentry{\bfseries\itshape MANN}

\draw[draw=none,fill=myblue,very thin] (axis cs:1.5,0) rectangle (axis cs:1.7,0.192857143);
\draw[draw=none,fill=myblue,very thin] (axis cs:2.5,0) rectangle (axis cs:2.7,0.533571429);
\draw[draw=none,fill=myblue,very thin] (axis cs:3.5,0) rectangle (axis cs:3.7,0.477857143);
\end{axis}

\end{tikzpicture}\hspace{-13pt}
\pgfplotsset{compat=1.11,
        /pgfplots/my ybar legend/.style={
        /pgfplots/legend image code/.code={%
        \draw[##1,/tikz/.cd,bar width=3pt,yshift=-0.2em,bar shift=0pt]
                plot coordinates {(0cm,0.8em)};},
},
}
\begin{tikzpicture}[scale=0.57]

\definecolor{dimgray85}{RGB}{85,85,85}
\definecolor{gainsboro229}{RGB}{229,229,229}
\definecolor{goldenrod1911910}{RGB}{191,191,0}
\definecolor{lightgray204}{RGB}{204,204,204}
\definecolor{gray}{RGB}{28,28,28}
\definecolor{lightgray}{RGB}{120,120,120}
\definecolor{myred}{HTML}{ff595e}
\definecolor{mygreen}{HTML}{8ac926}
\definecolor{myblue}{HTML}{4895ef}
\definecolor{myyellow}{HTML}{ffbe0b}

\begin{axis}[
width=0.9\linewidth,
height=0.55\linewidth,
axis background/.style={white},
axis line style={gray},
legend cell align={left},
legend columns=3,
legend style={fill opacity=0.8, draw opacity=1, text opacity=1, draw=none, at={(0.5,0.8)},anchor=south, fill=none,font=\footnotesize},
tick align=inside,
tick pos=left,
title style={font=\bfseries, at={(0.5,0.95)}},
title={w/o Real data},
x grid style={white},
xmajorgrids,
xmin=-0.2625, xmax=2.7625,
xtick style={color=dimgray85},
xtick={0.2,1.3,2.3},
xticklabel style={font=\bfseries\itshape},
xticklabels={Action QR,Smoothness,Naturalism},
y grid style={white},
ymajorgrids,
ymin=0, ymax=1.1,
ytick={0,0.2,0.4,0.6,0.8,1},
ylabel style={font=\bfseries\itshape},
ylabel=\textcolor{gray}{Score},
ytick style={color=dimgray85}
]
\addlegendimage{ybar,my ybar legend,draw=none,fill=myred,very thin}
\addlegendentry{\bfseries\itshape\underline{MarioNet}}
\addlegendimage{ybar,my ybar legend,draw=none,fill=mygreen,very thin}
\addlegendentry{\bfseries\itshape A2M}
\addlegendimage{ybar,my ybar legend,draw=none,fill=myblue,very thin}
\addlegendentry{\bfseries\itshape  MANN}

\draw[draw=none,fill=myred,very thin] (axis cs:-0.125,0) rectangle (axis cs:0.125,0.739);
\draw[draw=none,fill=myred,very thin] (axis cs:0.875,0) rectangle (axis cs:1.125,0.794285714);
\draw[draw=none,fill=myred,very thin] (axis cs:1.875,0) rectangle (axis cs:2.125,0.800714286);
\draw[draw=none,fill=mygreen,very thin] (axis cs:0.125,0) rectangle (axis cs:0.375,0.3455);
\draw[draw=none,fill=mygreen,very thin] (axis cs:1.125,0) rectangle (axis cs:1.375,0.462857143);
\draw[draw=none,fill=mygreen,very thin] (axis cs:2.125,0) rectangle (axis cs:2.375,0.449285714);
\draw[draw=none,fill=myblue,very thin] (axis cs:0.375,0) rectangle (axis cs:0.625,0.3964);
\draw[draw=none,fill=myblue,very thin] (axis cs:1.375,0) rectangle (axis cs:1.625,0.63);
\draw[draw=none,fill=myblue,very thin] (axis cs:2.375,0) rectangle (axis cs:2.625,0.567857143);
\end{axis}

\end{tikzpicture}\vspace{-10pt}
    \caption{Results of subject evaluation. ``A-1 Acc'' and ``A-2 Acc'' in the left figure indicate the action recognition of the first and the second actions respectively. Both figures plot average scores which range from $0 \sim 1$,  and $1$ represents the best. }\vspace{-10pt}
    \label{fig:res:us}
\end{figure}

Along with the above objective experiments, we also conduct a user study to evaluate the quality of actions purely generated by generative models.
The quality is judged from three aspects: action accuracy, transition smoothness, and naturalism.
The user study has two parts including 8 videos independently.
In the first part, each video contains two actions of human motions from the real data and those generated by MANN, A2M, and our method. 
We use the initial motion, action tags, and duration from the compared real data to generate motions. 
Based on the video contents ( Relative videos can be found in supplementary material), users are requested to verify whether the first and the second actions match to given action tags respectively. 
Then they give scores for transition smoothness and naturalism. 

The second part does \emph{not} include real motions. 
And every motion contains $3\sim 5$ actions generated by different methods with the same control signals.
In addition to the scores of transition smoothness and naturalism as in the previous part, users are asked to select the number of accurate actions corresponding to action tags for each motion.
Every user should finish both parts.
We collect 35 valid evaluation statistics from subjects varying in age, gender, and country.

The average results of the first part are presented in Fig.~\ref{fig:res:us} denoted as "W/ real data".  
"A1-Acc" and "A2-Acc" are the percentages of how many users believe actions match their tags for the first and the second actions respectively.
It can be found that our method has the closest accuracy performance to real data and a significantly higher score compared to MANN and A2M. 
Besides, the accuracy of the second action has an obvious decrease contrasted to the first one, which is consistent with the objective evaluation in Fig.~\ref{fig:res:mag_acc_fid}. 
We conclude this as the challenges of connection sparsity and error accumulation described in Sec.~\ref{subsec:hybrid}.
As for smoothness and naturalism, our method achieves the best performance among generative models but is still lower than real data.
Although no real motion is provided in the second part, we observe similar results that our method obtain much higher scores compared to MANN and A2M. 
Especially for Action QR, our method achieves \boldsymbol{$73.9\%$}, which is more than \boldsymbol{$30\%$} higher than the scores of A2M and MANN. 
This improvement is significantly larger than that in the objective study in Fig.~\ref{fig:res:mag_acc_fid}.

In summary, the subjective study demonstrates the consistent results with objective experiments that our method surpasses two compared motion synthesis methods in terms of action accuracy, transition smoothness, and naturalism. However, more effort should be made in future work that can help to decrease the gap between our synthetic motions and the real ones.

\subsubsection{Performance improvement of  ``SS'' and ``AR''}
In order to alleviate the phenomenon of BOP out-of-distribution sometimes caused by the common drifting problem in temporal auto-regressive models, we introduce ``Shadow Start'' denoted as SS, and ``Action Revision'' denoted as AR, to the motion synthesis pipeline. 
\begin{table}[!h]
    \centering
    \caption{Comparison of SS and AR. }
    \label{tab:res:hybrid}\vspace{-10pt}
    \centering
{\def\arraystretch{1}\tabcolsep=1.5em 
\begin{tabular}{@{}cccc@{}}
\toprule
Method          & Acc$\uparrow$     & $\Delta_{pose} \downarrow$ & $\Delta_{vel} \downarrow$ \\ 
\midrule
{\thename}         & 0.786  & \textbf{2.03}     & \textbf{0.40}    \\
{\thename}+SS      & \underline{0.819} & 4.57     & 0.47    \\
{$\text{\systemname}^*$}   & \textbf{0.853} & \underline{2.63}     & \underline{0.42}    \\
{\systemname}               & 0.896             & 0.265   &  0.401\\
\bottomrule
\end{tabular}
}

\end{table}
\begin{itemize}
[leftmargin=*]
    \item {\thename} is a pure generative pipeline with our model.
    \item{\thename}+SS indicates always using {\thename} with Shadow Start operation.
    \item {\systemname} is the hybrid method with a classifier to decide when to use the Shadow Start operation.
    \item {$\text{\systemname} ^*$} is full pipeline without using L2-normalization in SS step.
\end{itemize}

Quantitative results on \emph{overall} testing in Table~\ref{tab:res:hybrid} and Fig.~\ref{fig:res:hybrid_QR} shows the effectiveness of the two proposed schemes. 
``SS'' considers candidate BOPs in training data \emph{every} step to correct the drifting BOP.
As expected, it improves the quality of synthetic animation on Accuracy and FID while suffering from a more dramatic gap in smoothness. With the help of a recognition network, we could determine when is suitable to call the ''Shadow Start'', which extremely relieves undesired abrupt changes between sequential actions caused by projection operation. 
We also present a visual comparison example in Fig.~\ref{fig:res:ss}.(a).
It can be seen that directly generating (the 1st row) action with \thename~ leads to a ``frozen'' motion.
With our ``SS'', the ``walking backward'' action can be synthesized correctly.
The last row is a ``walking backward'' action we search from training set that has the most similar starting pose to the end one of the previous action.
By comparing these three start pose to the end pose of the previous action, we can find that ``SS'' provide a good compromise between the end poses of the previous action, which may be out-of-distribution sometimes, and start poses belong to the desired actions in the training set.

 

Since an action may fail only when both normal generation and re-generation using a projected BOP do not work at the same time, we get the best Accuracy by \boldsymbol{$85.3\%$} for over 14000 actions in total.
To conclude, recognition-based ``Action Revision''  makes {\systemname} more robust in high-quality motion generation while providing comparable performance in terms of Transition Smoothness compared to the pure generative pipeline.

\begin{figure}[!h]
  \centering
  
\pgfplotsset{
compat=1.11,
legend image code/.code={
\draw[mark repeat=2,mark phase=2]
plot coordinates {
(0cm,0cm)
(0.3cm,0cm)        
(0.2cm,0cm)         
};%
}
}
\begin{tikzpicture}[scale=1.0]

\definecolor{dimgray85}{RGB}{85,85,85}
\definecolor{gainsboro229}{RGB}{229,229,229}
\definecolor{gray}{RGB}{28,28,28}
\definecolor{lightgray}{RGB}{120,120,120}
\definecolor{myred}{HTML}{ff595e}
\definecolor{mygreen}{HTML}{8ac926}
\definecolor{myblue}{HTML}{4895ef}
\definecolor{myyellow}{HTML}{ffbe0b}
\pgfmathsetlengthmacro\MajorTickLength{
      \pgfkeysvalueof{/pgfplots/major tick length} * 0.5
    }
\begin{axis}[
width=0.55\linewidth,
axis background/.style={fill=white},
axis line style={gray},
legend cell align={left},
legend style={
  fill opacity=0.8,
  draw opacity=1,
  text opacity=1,
  at={(0.4,0.0)},
  anchor=south,
  draw=none,
  fill=none,
  font=\tiny
},
tick align=inside,
tick pos=left,
major tick length=\MajorTickLength,
title style={font=\footnotesize, at={(0.5,0.95)}},
x grid style={gray!10},
xlabel style={at={(axis description cs:0.5,-0.1)},font=\footnotesize\itshape},
xlabel=\textcolor{gray}{Number of actions},
xmajorgrids,
xmin=1, xmax=20,
xtick style={color=dimgray85},
y grid style={gray!10},
ylabel style={at={(axis description cs:-0.15,.5)},font=\footnotesize\itshape},
ylabel=\textcolor{gray}{Action QR},
ymajorgrids,
ymin=0.6, ymax=1,
ytick style={color=dimgray85},
tick label style={font=\footnotesize}
]
\addplot [ line width=1.0pt, myred]
table {%
1 0.945
2 0.899
3 0.874333333333333
4 0.8605
5 0.8478
6 0.840166666666667
7 0.837571428571429
8 0.83025
9 0.824333333333333
10 0.82
11 0.817909090909091
12 0.812166666666667
13 0.808384615384616
14 0.804
15 0.8022
16 0.800875
17 0.798470588235294
18 0.797388888888889
19 0.796157894736842
20 0.7937
};
\addlegendentry{MARIONET}
\addplot [ line width=1.0pt, mygreen]
table {%
1 0.945
2 0.9045
3 0.881666666666667
4 0.87075
5 0.8664
6 0.8595
7 0.855285714285714
8 0.853875
9 0.852222222222222
10 0.8489
11 0.846363636363636
12 0.843333333333333
13 0.840769230769231
14 0.837642857142857
15 0.8346
16 0.8314375
17 0.830352941176471
18 0.829666666666667
19 0.82721052631579
20 0.82575
};
\addlegendentry{MARIONET+SS}
\addplot [ line width=1.0pt, myblue]
table {%
1 0.945
2 0.9235
3 0.906333333333333
4 0.90125
5 0.8942
6 0.889333333333333
7 0.888857142857143
8 0.888
9 0.884888888888889
10 0.8815
11 0.878909090909091
12 0.875916666666667
13 0.873461538461539
14 0.871
15 0.869133333333333
16 0.865875
17 0.863529411764706
18 0.8615
19 0.859473684210526
20 0.85725
};
\addlegendentry{NEURAL MARIONETTE$^*$}

\addplot [ line width=1.0pt, myyellow]
table {%
1	0.936
2	0.924
3	0.915
4	0.912
5	0.898
6	0.891
7	0.91
8	0.894
9	0.906
10	0.894
11	0.894
12	0.889
13	0.903
14	0.877
15	0.885
16	0.872
17	0.88
18	0.897
19	0.894
20	0.894

};
\addlegendentry{NEURAL MARIONETTE}

\addplot [ line width=0.5pt, gray, dashed, forget plot]
table {%
2 0.6
2 1
};

\end{axis}

\end{tikzpicture}\hspace{-13pt}
\begin{tikzpicture}[scale=1.0]
\pgfmathsetlengthmacro\MajorTickLength{
      \pgfkeysvalueof{/pgfplots/major tick length} * 0.5
    }
\definecolor{dimgray85}{RGB}{85,85,85}
\definecolor{gainsboro229}{RGB}{229,229,229}
\definecolor{gray}{RGB}{28,28,28}
\definecolor{lightgray}{RGB}{120,120,120}
\definecolor{myred}{HTML}{ff595e}
\definecolor{mygreen}{HTML}{8ac926}
\definecolor{myblue}{HTML}{4895ef}
\definecolor{myyellow}{HTML}{ffbe0b}

\begin{axis}[
width=0.55\linewidth,
axis background/.style={fill=white},
axis line style={gray},
legend cell align={left},
legend columns=4,
legend style={
  fill opacity=0.8,
  draw opacity=1,
  text opacity=1,
  at={(0.5,0.0)},
  anchor=south,
  draw=none,
  fill=none
},
tick align=inside,
tick pos=left,
major tick length=\MajorTickLength,
title style={font=\footnotesize, at={(0.5,0.95)}},
x grid style={gray!10},
xlabel style={at={(axis description cs:0.5,-0.1)},font=\footnotesize\itshape},
xlabel=\textcolor{gray}{Number of actions},
xmajorgrids,
xmin=1, xmax=20,
xtick style={color=dimgray85},
y grid style={gray!10},
ylabel style={at={(axis description cs:-0.15,.5)},font=\footnotesize\itshape},
ylabel=\textcolor{gray}{Motion QR},
ymajorgrids,
ymin=0, ymax=1,
ytick style={color=dimgray85},
tick label style={font=\footnotesize}
]
\addplot [line width=1.0pt, myred ]
table {%
1 0.945
2 0.827
3 0.724
4 0.661
5 0.572
6 0.525
7 0.475
8 0.436
9 0.391
10 0.365
11 0.343
12 0.297
13 0.266
14 0.236
15 0.219
16 0.191
17 0.171
18 0.162
19 0.145
20 0.127
};
\addplot [ line width=1.0pt, mygreen ]
table {%
1 0.945
2 0.831
3 0.742
4 0.668
5 0.61
6 0.56
7 0.501
8 0.463
9 0.428
10 0.394
11 0.349
12 0.318
13 0.287
14 0.257
15 0.24
16 0.207
17 0.192
18 0.178
19 0.161
20 0.145
};
\addplot [line width = 1.0pt, myblue ]
table {%
1 0.945
2 0.868
3 0.792
4 0.747
5 0.694
6 0.653
7 0.622
8 0.601
9 0.562
10 0.524
11 0.488
12 0.454
13 0.432
14 0.404
15 0.384
16 0.356
17 0.336
18 0.321
19 0.3
20 0.274
};

\addplot [ line width=1.0pt, myyellow]
table {%
1	0.936
2	0.88
3	0.832
4	0.787
5	0.745
6	0.712
7	0.678
8	0.645
9	0.622
10	0.597
11	0.575
12	0.547
13	0.53
14	0.507
15	0.484
16	0.457
17	0.44
18	0.432
19	0.41
20	0.399
};

\addplot [line width=0.5pt, gray, dashed, forget plot]
table {%
2 0
2 1
};
\end{axis}

\end{tikzpicture}
    \vspace{-20pt}
    \caption{Action QR (left) and Motion QR (Right) of Our variants. 
    }
    \label{fig:res:hybrid_QR}
\end{figure}\vspace{-10pt}
\begin{figure}[!h]
\vspace{-5pt}
 \subfloat[Action synthesis w/o and w/ SS.]{ \includegraphics[width=0.48\linewidth]{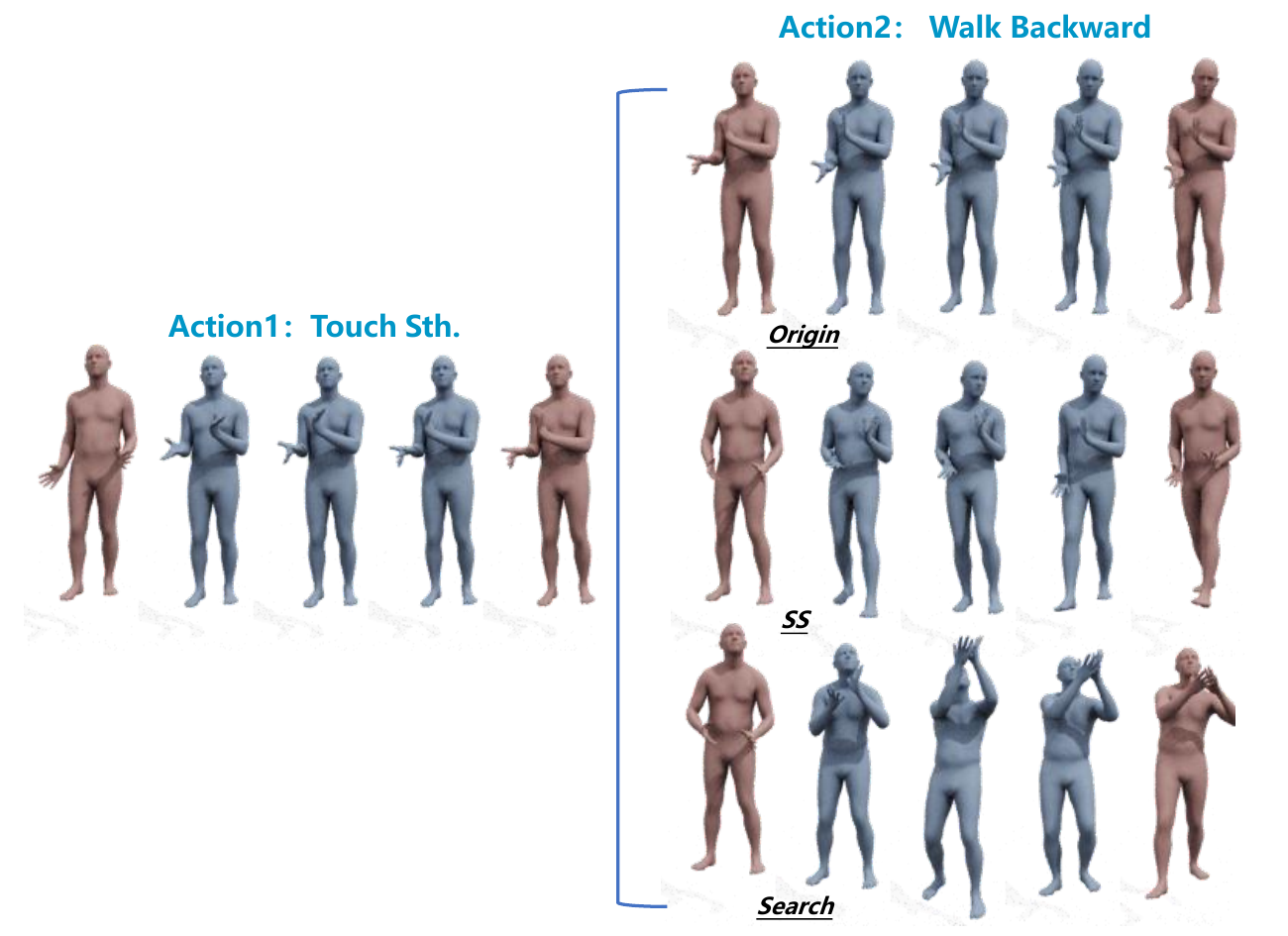}\vspace{-10pt}} 
 \hfill
 \subfloat[Failure case]{ \includegraphics[width=0.48\linewidth]{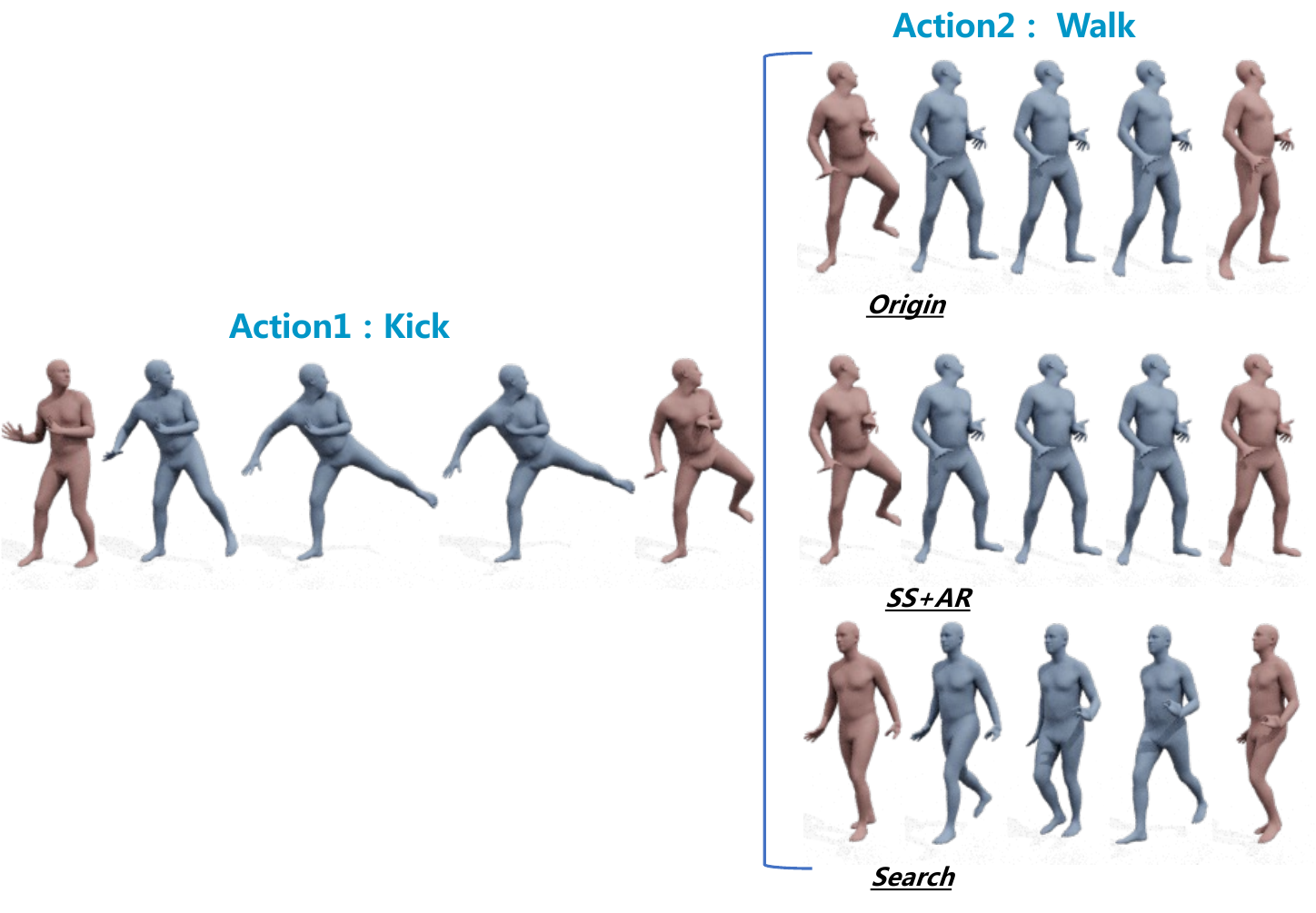}\vspace{-10pt}}
 
  \caption{\textbf{(a)} Action synthesis w/o and w/ SS. The 1st and 2nd rows are action generated w/o and w/ SS, respectively. W/o SS, motion in the 1st row is ``frozen'', while w/ SS, the second row shows the synthesized action is correct. The last row is an action with the most similar starting pose to the end pose of the previous action. \textbf{(b)}. Failure case From  ~\systemname. In this case, our Action Revision did not recognize incorrect action to trigger off ``SS'' operation. }\vspace{-5pt}
 \label{fig:res:ss}
\end{figure}

%
\subsubsection{Failure case}

 
%
From the above comparison, it can be seen that our \systemname~ achieves significantly higher performance. 
It has to be pointed out that, even with ``action revision'' and ``shadow start'', our method may fail to generate correct actions, especially when dealing with unconstrained user inputs.
Failure cases usually appear when the end pose of the previous action is significantly different from the start pose of the current action in training sets. 
Fig.~\ref{fig:res:ss}. (b) presents an example to illustrate how the multi-action synthesis failed. 

The left part of Fig.~\ref{fig:res:ss}. (b) is a ``kick'' action while the first and second rows of the right part are "walk" actions generated by the pure generative pipeline and ~{\systemname} respectively. It can be found that they are the same action. That is because our AR did not recognize this unexpected action to trigger off ``SS'' scheme. 
The last row is another action with the "walk" tag which is retrieved from the training dataset with the closest starting pose to the end pose of the previous action. 
This starting pose has a significant difference from the previous end one, making our {\thename}~ hard to generate the correct action.
Besides, the unreasonable action sequence, such as ``sit down--run'' (``run'' should happen after a ``stand up'' action), extremely short duration and wired trajectory may also lead to a failure motion synthesis for ~\systemname.

%

   

    

\section{ Interactive Motion Synthesis}\label{sec:system}

\begin{figure}[!h]
    \centering
    
   \subfloat[Trajectory Drawing]
   { \includegraphics[height =0.4\linewidth, width = 0.45\linewidth]{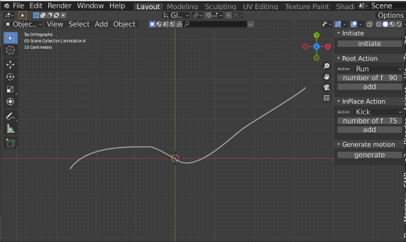}}
   \hfill
    \subfloat[ Per-action annotation]{
    \includegraphics[height =0.4\linewidth,width = 0.45\linewidth]{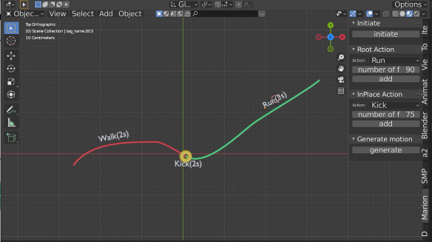}}
    \caption{Interactive steps of multi-action synthesis.  }
    \label{fig:steps}
\end{figure}

\begin{figure}[!h]
    \centering
    \includegraphics[height=0.6\linewidth,width=1.0\linewidth]{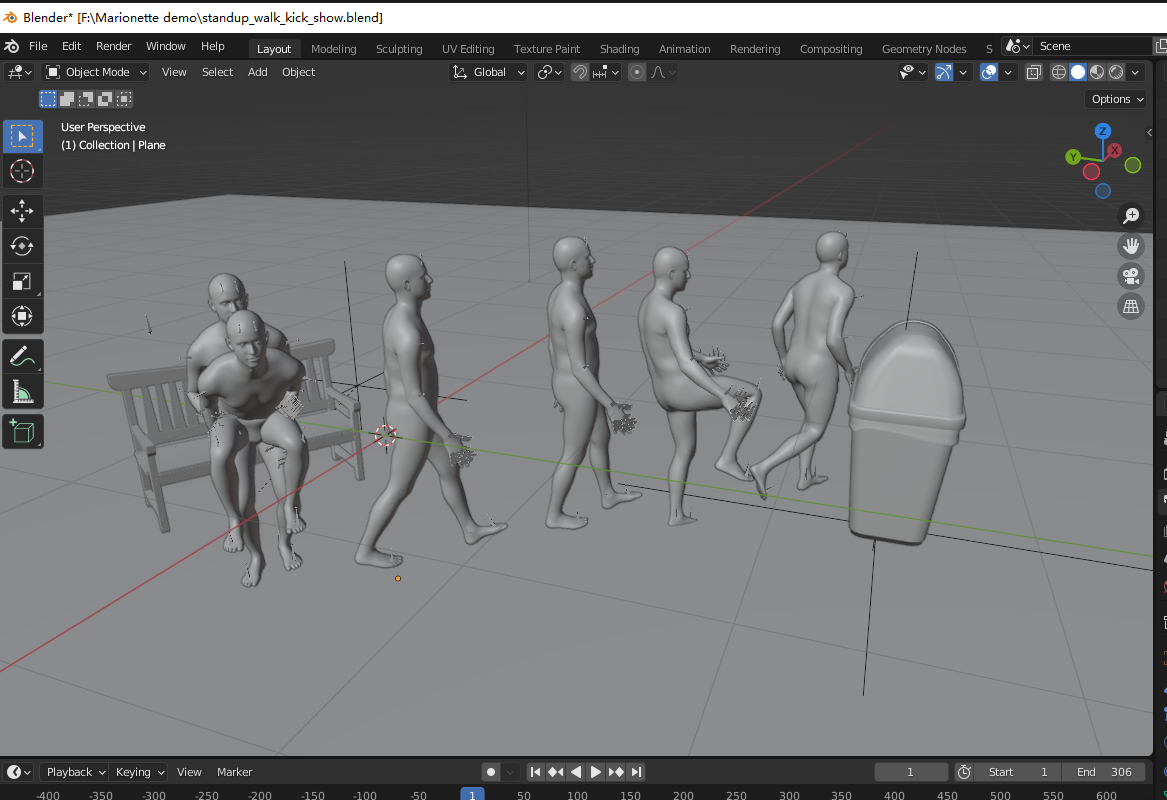}\vspace{-8pt}
    \caption{A user case of our interactive motion synthesis GUI. It demonstrates a man \emph{stands up}, then \emph{walks} to a trash bin then \emph{kick} it, and finally \emph{runs} away. }\vspace{-10pt}
    \label{fig:system:ui_case}
\end{figure}

In this section, we demonstrate that by using our {\systemname}, multi-action can be controllably generated in an interactive fashion.
We develop an interactive multi-action system as a blender add-on and the parametric human model is from SMPLX~\cite{SMPL-X:2019}.
Besides our \systemname, a simple spherical linear interpolation~\cite{shoemake1985animating} is used to further smooth the transition when ``Shadow Start'' is used.
%
%
Users can generate a multi-action motion by first plotting a free hand curve as Fig~\ref{fig:steps}.(a), then annotating this curve with multiple action tags, and duration as Fig.~\ref{fig:steps}.(b).
After finishing per-action annotating, the user can click the ``generating'' button, then our \systemname~ starts to generate actions and finally displays them on the screen.
We present a human motion generated with our interactive system in Fig~\ref{fig:system:ui_case}.

%

\section{Conclusion and  limitation}
In conclusion, we propose a new scheme, {\systemname}, for the multi-action human motion synthesis task. 
{\thename}, the core part of our method, can fully utilize action tag and contextual information to generate high-quality single-action motions with smooth transitions which can be later stitched together to produce a much longer multi-action motion. 
Experiments on both conditional single-action and multi-action generation tasks demonstrate that our {\thename} and {\systemname} achieve superior motion synthesis performance in terms of action accuracy, naturalism, and transition smoothness over multiple baseline methods. 
We further demonstrate that our method can be used for human animation creation in an interactive way via a Blender add-on integrated with our  {\systemname}.

Although our method has achieved state-of-the-art performance among generative models, the result of subject evaluation suggests that there is still a gap between the real human motions and our synthesized motions in terms of {Realism} and {Transition Smoothness}. 
Another limitation of our {\systemname} is that it does not support interactive frame-editing during generation which limits the usage in real animation creation where more strict requirements have to be satisfied.
Though the whole generated motion can be edited afterward, it would be more convenient for technical artists to generate the desired motion if frame editing is available during the current interactive workflow.
Therefore, in the future, we would like to investigate more powerful techniques to narrow the gap between synthetic motions and real ones. 
Besides, we also notice that a foot-sliding problem may happen when generating 
root motion actions. 
This is caused mainly by two reasons. First, the training data are from different people with different bone sizes. Foot-sliding naturally exists when transferring motion to another character. Second, we do not consider physical constrain in our model. we leave this problem in our future work.


%



\ifCLASSOPTIONcaptionsoff
  \newpage
\fi



%


\bibliographystyle{IEEEtran}
\bibliography{newReference}





\end{document}


%
\title{Supplementary for \systemname: A Transformer-based
Multi-action Human Motion Synthesis System}
\author{Weiqiang~Wang \IEEEauthorrefmark{2},
         Xuefei~Zhe \IEEEauthorrefmark{2},  
         Qiuhong~Ke, 
         Di~Kang, 
         Tingguang~Li, 
        Ruizhi~Chen, 
        and~Linchao~Bao \IEEEauthorrefmark{1}
\IEEEcompsocitemizethanks{
\IEEEcompsocthanksitem W. Wang and Q. Ke are with the Faculty of Information Technology, Monash University, Melbourne 3168, Australia. 
\protect\\ 
E-mail: weiqking@gmail.com, qiuhong.ke@monash.edu.

\IEEEcompsocthanksitem X. Zhe, D. Kang and L. Bao are with Tencent AI Lab, Shenzhen 518054, China. 
\protect\\ 
E-mail: \{zhexuefei,~di.kang\}@outlook.com, linchaobao@gmail.com.

\IEEEcompsocthanksitem T. Li is with Tencent Robotics X, Shenzhen 518054, China. 
\protect\\ 
E-mail: tgli0809@gmail.com.

\IEEEcompsocthanksitem
R. Chen is with the State Key Laboratory of Information Engineering in Surveying, Mapping, and Remote Sensing, Wuhan University, Wuhan 430079, China. 
\protect\\ 
E-mail: ruizhi.chen@whu.edu.cn.

\IEEEcompsocthanksitem W. Wang did the work during his internship at Tencent AI Lab.\\
\IEEEauthorblockA{\IEEEauthorrefmark{2} W. Wang and X. Zhe contributed equally to this work. }\\
 \IEEEauthorblockA{\IEEEauthorrefmark{1} L. Bao is the corresponding author. }
}
}

\maketitle

\IEEEdisplaynontitleabstractindextext

%
\IEEEpeerreviewmaketitle


%

%
\section{Terminology}
In this section, we briefly provide some terminologies used in our paper for better understanding.
\subsection{Motion and Action}
We define a \emph{motion} as a sequence of human poses with different movements, which can be long and complicated. 
As a comparison, an \emph{action} is a relatively short motion clip, which only contains a single type of movement pattern.
Therefore, a motion can be decomposed into multiple sequential actions. 
For example, a motion sequence that has a human walking for a few seconds and then running can be decomposed into two actions, "Walk" and "Run". 

\subsection{Action Description} 
We first define $\mathcal{A}$ as a collection of action \emph{tags} that are as independent of each other as possible.
As discussed above, an action can then be described by tags from $\mathcal{A}$, i.e., $\Scale[0.95]{A = \mybrac{a_{1}, \cdots, a_{k}}}$, where $\Scale[0.95]{a_{i}\in\mathcal{A}}, \forall i=1\cdots k$. 
We call this action has \emph{Tag Annotation} $A$.
Note that, multi-tag annotation happens only for non-intention actions 

\subsection{Action Representation}
An action $M$ can be represented by a sequence of human poses, i.e., $\Scale[0.95]{M = \mybrac{P_1, \cdots, P_t, \cdots, P_T}}$, where $\Scale[0.95]{P_t}$ denotes a single pose at $t$-th frame and $T$ is the total number of frames in the clip. 
A human pose $P_t$ can be characterized by a \emph{root translation} $D_t\in\mathbf{R}^{3}$ and the \emph{rotations} of all the joints $R_t\in\mathbf{R}^{n_J\times 6}$, where $n_J$ is the number of joints and the continuous $6$D rotation representation is used~\cite{zhou2019continuity}.
Given a pose $\Scale[0.95]{P_t = \big(\, R_t, D_t \,\big)}$ and certain body shape parameter, we can recover the \emph{body mesh} using SMPL~\cite{smpl} or SMPLX~\cite{SMPL-X:2019}.

\subsection{Action Trajectory}
The trajectory $U$ of an action $M$ can be described by \emph{root translation} $D_t (t=1,\cdots, T)$. We denote this as a function, $F: [1,\cdots, T] \rightarrow \mathbf{R}^2$, which assigns a translation vector in the ground plane to a specified frame.
Thus, one way to directly manipulate the trajectory of the synthesized motion is by drawing a curve on the ground in the world coordinate system and then sampling discretely. More details are provided in Sec.~\ref{appendix:preprocess}.


\section{Dataset}
\subsection{{\dtname} Dataset}\label{appendix:dt}


In this part, we show more details about the {\dtname}. 
To clean the annotations of the collected motion data, we propose an intention-first tagging protocol, including a simple tagging pipeline with consist of several tagging rules, which instruct how to compose tags to consistently and thoroughly describe an action. 
Specifically, we first annotate those actions with a clear intention such as "punch" and "sit down". These actions are tagged according to their final goals. For example, if an action demonstrates a man \emph{walks} for \emph{moving something}, we will tag this action with "moving something" because his final goal is to move something and walking is helping to achieve this goal.
If an action does not have a clear intention, we then use the corresponding body state (including "stand", "bend", and "sit") and body part (including "head", "arm", "waist", and "leg") movement to describe it.
Note that only non-intention actions are allowed to be labelled with multiple tags.
Our action tags have simple and unambiguous definitions, and they are as independent as possible from each other, greatly simplifying the action annotation process. 

\begin{figure}[!ht]
    \centering
\begin{tikzpicture}

\begin{axis}[
width=0.85\linewidth,
height=0.85\linewidth,
colorbar,
colorbar style={ylabel={}},
colormap={mymap}{[1pt]
  rgb(0pt)=(0.968627450980392,0.988235294117647,0.941176470588235);
  rgb(1pt)=(0.87843137254902,0.952941176470588,0.858823529411765);
  rgb(2pt)=(0.8,0.92156862745098,0.772549019607843);
  rgb(3pt)=(0.658823529411765,0.866666666666667,0.709803921568627);
  rgb(4pt)=(0.482352941176471,0.8,0.768627450980392);
  rgb(5pt)=(0.305882352941176,0.701960784313725,0.827450980392157);
  rgb(6pt)=(0.168627450980392,0.549019607843137,0.745098039215686);
  rgb(7pt)=(0.0313725490196078,0.407843137254902,0.674509803921569);
  rgb(8pt)=(0.0313725490196078,0.250980392156863,0.505882352941176)
},
point meta max=100,
point meta min=0,
tick align=outside,
tick pos=left,
title={Tags Adjacent Matrix},
x grid style={white!69.0196078431373!black},
xlabel={Tag ID},
xmin=0, xmax=31,
xtick style={color=black},
xtick={0.5,2.5,4.5,6.5,8.5,10.5,12.5,14.5,16.5,18.5,20.5,22.5,24.5,26.5,28.5,30.5},
xticklabel style={font=\footnotesize},
xticklabels={0,2,4,6,8,10,12,14,16,18,20,22,24,26,28,30},
y grid style={white!69.0196078431373!black},
ylabel={Tag ID},
ymin=0, ymax=31,
ytick style={color=black},
ytick={0.5,2.5,4.5,6.5,8.5,10.5,12.5,14.5,16.5,18.5,20.5,22.5,24.5,26.5,28.5,30.5},
yticklabel style={rotate=90.0,font=\footnotesize},
yticklabels={0,2,4,6,8,10,12,14,16,18,20,22,24,26,28,30}
]
\addplot graphics [includegraphics cmd=\pgfimage,xmin=0, xmax=31, ymin=0, ymax=31] {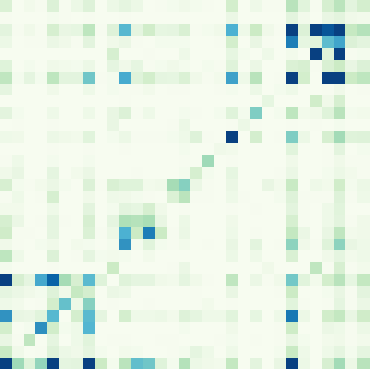};
\end{axis}

\end{tikzpicture}
    \vspace{-25pt}
    \caption{We report the number of action pairs that are adjacent to each other between every two tags via a heatmap.}
    \label{fig:append:babel_adj_new}
\end{figure}

The statistic and type of every action tag used in {\dtname} are presented in Table ~\ref{appdendix:action_tag_tab}. 
Besides, we calculate the action pairs between every two tags and show this by the Transition Adjacent Matrix via a heatmap in Fig.~\ref{fig:append:babel_adj_new}.

\begin{table}[!ht]
    \centering
    \caption{Action Tags for {\dtname}. Note that, five action types --- "Intention", "Root", "In-place", "Status", "Body parts" are correspondingly abbreviated as "I", "R", "P", "S", "B" for simplification.}\vspace{-10pt}
    \label{appdendix:action_tag_tab}
    \centering
{\def\arraystretch{1}\tabcolsep=0.5mm
\begin{tabular}{cccc}
\toprule
Tag ID          & Tag Type                  &  Tag Name                & Segment Number \\ \midrule
0               & I \& R                    & Walk                     & 1370           \\
1               & I \& R                    & Walk Backward            & 160            \\
2               & I \& R                    & Walk Sideway             & 105            \\
3               & I \& R                    & Walk Upward/Downward     & 272            \\
4               & I \& R                    & Step                     & 536            \\
5               & I \& R                    & Run                      & 198            \\
6               & I \& R                    & Jump                     & 168            \\
7               & I \& P                    & Turn                     & 1063           \\
8               & I \& P                    & Sit Down                 & 86             \\
9               & I \& P                    & Stand Up                 & 123            \\
10              & I \& P                    & Using Object             & 227            \\
11              & I \& P                    & Take Something Up        & 375            \\
12              & I \& P                    & Place Something Down     & 261            \\
13              & I \& P                    & Move Something           & 102            \\
14              & I \& P                    & Throw                    & 143            \\
15              & I \& P                    & Touch Something          & 301            \\
16              & I \& P                    & Kick                     & 91             \\
17              & I \& P                    & Punch                    & 69             \\
18              & I \& P                    & Wave Hand                & 29             \\
19              & I \& P                    & Stretch                  & 410            \\
20              & I \& P                    & Crossing Limbs           & 33             \\
21              & I \& P                    & Touch Body Part          & 200            \\
22              & I \& P                    & Using Object and Sit     & 39             \\
23              & I \& P                    & Step and Touch Body Part & 54             \\
24              & S                         & Stand                    & 754            \\
25              & S                         & Bend                     & 149            \\
26              & S                         & Sit                      & 325            \\
27              & B                         & Head Movement            & 199            \\
28              & B                         & Arm Movement             & 859            \\
29              & B                         & Waist Movement           & 30             \\
30              & B                         & Leg Movement             & 193            \\ \bottomrule
\end{tabular}}
\end{table}

\section{Implementation}
We implemented {\thename} with Pytorch, and {\systemname} as an add-on in Blender. All motion sequences are firstly downsampled to 30 Hz, and the initial motion provided in each contextual information is fixed to 6 frames, while tag annotations of {\dtname} are saved in the same format as BABEL.
The detailed settings for {\thename} is shown in Table~\ref{tab:network}. And we trained it using Adam~\cite{Kingma2015AdamAM} with a batch size of 30 and a learning rate of 0.0001, for 1000 epochs where Activate epoch is 250 and Accomplish epoch is 750. Besides, we empirically set different weights of training loss with  $\lambda_{KL}=10^{-8},\lambda_D=0.5,\lambda_R=1,\lambda_J=2,\lambda_V=2$. 
When the scheduled sampling is activated,$\lambda_1=0.1$ and $\lambda_2=0.9$; $\lambda_1=1$ and $\lambda_2=0$ otherwise.

\emph{Scheduled Sampling for Transformer.} There is a severe train-test discrepancy called "Exposure Bias" in the teacher-forcing training scheme for sequence-related generation~\cite{ranzato2015sequence,zhang2019bridging,huang2020dance}. 
%
The Scheduled Sampling strategy is proposed to alleviate this dilemma in sequence generation~\cite{bengio2015scheduled}.
It is relatively easy to adapt Scheduled Sampling to motion generation for models with RNN architecture~\cite{huang2020dance} since they generate a motion sequence frame by frame during the training phase.
However, Transformer processes the input and outputs the whole sequence in parallel by the attention-mask mechanism during training, making it hard to take a previously predicted pose as input in a frame-by-frame fashion~\cite{Mihaylova2019SS,li2021scheduled}. 
\begin{figure}[!h]
    \centering
    \begin{overpic}[trim=0cm 27.2cm 14.1cm 0cm,clip,width=0.9\linewidth,grid=false]{fig2_training.pdf}
    \end{overpic}\vspace{-12pt}
    \caption{\textbf{Scheduled Sampling Strategy}. \emph{Left}: the Motion Decoder takes the ground-truth poses $\{P_t\}_{t=0\cdots T-1}$ as input and output the synthesized poses $\{P'_{t}\}_{t=1\cdots T}$. \emph{Right}: $\{P_t\}$ and $\{P'_t\}$ are \emph{mixed} and feed into the Motion Decoder again to synthesize new poses $\{P''_t\}_{t=1\cdots T}$. Both $P'_t$ and $P''_t$ are used for loss construction. For simplicity, we use an orange/blue box to represent the pose encoding/decoding FC layer.  
    } \vspace{-10pt}
    \label{fig:mtd:sampling}
\end{figure}

In this part, we demonstrate how we adapt the Scheduled Sampling to motion generation with Transformer. %
Specifically, as illustrated in Fig.~\ref{fig:mtd:sampling}, two weights-sharing Motion Decoders with Transformer Decoder are involved. 
The first Decoder (left) takes the ground truth motion sequence $P = [P_0, P_1, \cdots, P_{T-1}]$ as input, and outputs the predicted motion sequence  $P' = [ P'_1, P'_2, \cdots, P'_{T}]$ that is one frame later than the input. 
Then the predicted motion $P'$ is \emph{mixed} with the ground truth $P = [P_0, P_1, \cdots, P_{T-1}]$ which we call the mixed motion $P^{\text{mix}}$. 
Note that, the ratio ($\xi$) of poses from the predicted motion can vary in $P^{\text{mix}}$ during the training. 
The mixed motion $P^{\text{mix}}$ is then fed into the second Decoder (right), which outputs a new predicted motion sequence $P''= [P''_1, P''_, \cdots, P''_T]$. 
The reconstruction loss takes both $P'$ and $P''$ into consideration when this strategy is adopted, which helps to bridge the gap between training and inference.
Note that, to avoid error accumulation of these predicted high-biased poses, our Scheduled Sampling strategy is activated only when the predicted motion $P'$ is close enough to the input ground-truth $P$.
During the training, the ratio $\xi$ is increased linearly from 0\% (i.e., $P^{\text{mix}} = P$) to 100\% (i.e., $P^{\text{mix}} = P'$) between the \emph{Activate} and \emph{Accomplish} epoch.
%
This strategy alleviates the train-test discrepancy, making our generated motion sequence both fluent and continuous at the testing phase.

 We will release our code and {\dtname} to reproduce all the results.

\begin{table*}[!h]
    \centering
    \caption{Network architecture for {\thename}.}
    \label{tab:network}
    \vspace{-8pt}
\centering
{\def\arraystretch{1}\tabcolsep=0.4em 
\begin{tabular}{|cc|cccc|cc|cc|}
\hline
\textbf{}                                                    &                                 & \multicolumn{4}{c|}{}                                                                 & \multicolumn{2}{c|}{\textbf{Time Unrolling}}   & \multicolumn{2}{c|}{{\color[HTML]{333333} }}                                          \\
                                                             & \textbf{}                       & \multicolumn{4}{c|}{\multirow{-2}{*}{\textbf{Condition Encder}}}                      & \textbf{Trajectory} & \textbf{Unrolling}       & \multicolumn{2}{c|}{\multirow{-2}{*}{{\color[HTML]{333333} \textbf{Motion Decoder}}}} \\ \hline
\multicolumn{1}{|c|}{}                                       & {\color[HTML]{333333} d\_model} & \multicolumn{4}{c|}{256}                                                              & 32                  & 256                      & \multicolumn{2}{c|}{{\color[HTML]{333333} 256}}                                       \\
\multicolumn{1}{|c|}{}                                       & n\_head                         & \multicolumn{4}{c|}{8}                                                                & 1                   & {\color[HTML]{333333} 8} & \multicolumn{2}{c|}{8}                                                                \\
\multicolumn{1}{|c|}{}                                       & dim\_feedforward                & \multicolumn{4}{c|}{1024}                                                             & 64                  & 1024                     & \multicolumn{2}{c|}{1024}                                                             \\
\multicolumn{1}{|c|}{}                                       & num\_layers                     & \multicolumn{4}{c|}{2}                                                                & 1                   & 2                        & \multicolumn{2}{c|}{6}                                                                \\
\multicolumn{1}{|c|}{}                                       & dropout                         & \multicolumn{4}{c|}{0.1}                                                              & 0.001               & 0.01                     & \multicolumn{2}{c|}{0.1}                                                              \\
\multicolumn{1}{|c|}{\multirow{-6}{*}{\textbf{Transformer}}} & activation                      & \multicolumn{4}{c|}{gelu}                                                             & relu                & relu                     & \multicolumn{2}{c|}{gelu}                                                             \\ \hline
\multicolumn{1}{|c|}{}                                       &                                 & \textbf{tag\_embd} & \textbf{pose\_encode} & \textbf{mean\_embd} & \textbf{std\_embd} & \textbf{traj\_embd} & \textbf{unroll\_embd}    & \textbf{pose\_encode}                     & \textbf{pose\_decode}                     \\ \cline{3-10} 
\multicolumn{1}{|c|}{}                                       & in\_size                        & 31                 & 135                   & 256                 & 256                & 2                   & 256                      & 135                                       & 256                                       \\
\multicolumn{1}{|c|}{}                                       & out\_size                       & 256                & 256                   & 256                 & 256                & 32                  & 224                      & 256                                       & 135                                       \\
\multicolumn{1}{|c|}{\multirow{-4}{*}{\textbf{FC}}}          & activation                      & relu               & relu                  & relu                & relu               & relu                & relu                     & relu                                      & relu                                      \\ \hline
\end{tabular}
}
\end{table*}

%
\section{More Experiments details}
\subsection{Details of baselines and their adaption}
\begin{itemize}
[leftmargin=*]
\item \textbf{MANN}~\cite{zhang2018mann} is a mixture-of-expert (MoE) based method for motion generation, consisting of $K$ expert network(s) and a gating network to output the blend weights of the experts.
Although MANN is designed for quadruped characters, its learned blending strategy takes complicated dynamic states into consideration automatically and outperforms PFNN~\cite{PFNN} which uses both a predefined blending strategy and manually labeled phase labels. 

\item \textbf{Action2motion (A2M)}~\cite{guo2020action2motion} is a temporal variational auto-encoder (VAE) which can generate 3D human motion based on the input action tag. From the view of input/output,  A2M is the closest method to ours.
For a fair comparison, we made two adjustments to A2M including (1) replacing lie algebra representation of the joint rotations with 6D continuous representation~\cite{zhou2019continuity}; (2) providing A2M with the same initial pose as our setting rather than a rest pose.
\end{itemize}


\subsection{Construction for Multi-action Evaluation Set}
We first use a trained action classifier to eliminate those test samples with lower confidence scores than \boldsymbol{$50\%$} to obtain action candidates set of \boldsymbol{$764$} high-quality action clips.
In the overall testing set, we firstly use any action with tag $A_0$ as the first action.
Then a next tag $A_1$ is randomly selected from all tags except ones that are not connected by the current tag $A_0$.
Finally, the next action is randomly chosen from actions with $A_1$ in the action candidates set.
A multi-action test sample is formed by \boldsymbol{$20$} actions selected in this way.
By going through the action candidates set, we obtain an overall testing dataset with \boldsymbol{$764$} multi-action samples.

As for sufficient testing, we consider building multi-action animation with sufficient connections between actions.
The difference from the overall testing is how the next action tag is selected.
We restrict the next candidate action tag derived from the following tags:
\begin{enumerate}
    \item  the top 4 tags connected by current action $A_0$ in transition matrix; 
    \item  the current action tag $A_0$
    \item  the next action tag of the current action sample 
\end{enumerate}

Then next tag is randomly selected from these candidate tags. 
We generated the same number of sufficient multi-action testing samples as the overall.
Under such a setting, we expect to study the performance of training data with sufficient action connections.

\subsection{Pure generative Algorithm}
We present the pure generative algorithm used in the multi-action experiment in Algo. ~\ref{algo:milti-action}, where function $\mathrm{G}$ is denoted for different generative models.

\begin{algorithm}[!h]
\SetAlgoLined
\SetKwInOut{Input}{Input}\SetKwInOut{Output}{Output}
\Input{$\{A_i,T_i,U_i\}_N $,  $\{P_t\}_{t=-x}^{t=-1}$}
\For{$i \gets 1$ \KwTo $N$}{
     $R_z ~ \text{and} ~ D_t \gets P_{-1}$ \;
    Normalized initial pose  $\{\,P^*_t\,\}_{t=-x}^{t=-1} \gets R_z^{-1} \{P_t\}_{t=-x}^{t=-1} - D_t$  \;
    Generate $i$-th action  $\{\,\tilde{P_t}\,\}_{t=1}^{t=T_i} \gets $ $\mathrm{G}(\{P^*\}_{t=-x}^{t=-1},A_i,A_{i+1},T_i,U_i)$\;
    Rotated i-th action $\{\,\hat{P_t}\,\}_{t=1}^{t=T_i} \gets  R_z \tilde{P_t} + D_t $ \;
    Update initial pose sequence  $\{\,P_t\,\}_{t=-x}^{t=-1} \gets 
    \{\, \hat{P_t}\,\}_{t=T_i-x+1}^{t=T_i}$ \;
    }
\caption{Pure generative multi-action pipeline}
\label{algo:milti-action}
\end{algorithm}

\subsection{Additional Results}
\label{appendix:res}
\begin{figure}[!ht]
    \centering
    \includegraphics[width=0.5\textwidth]{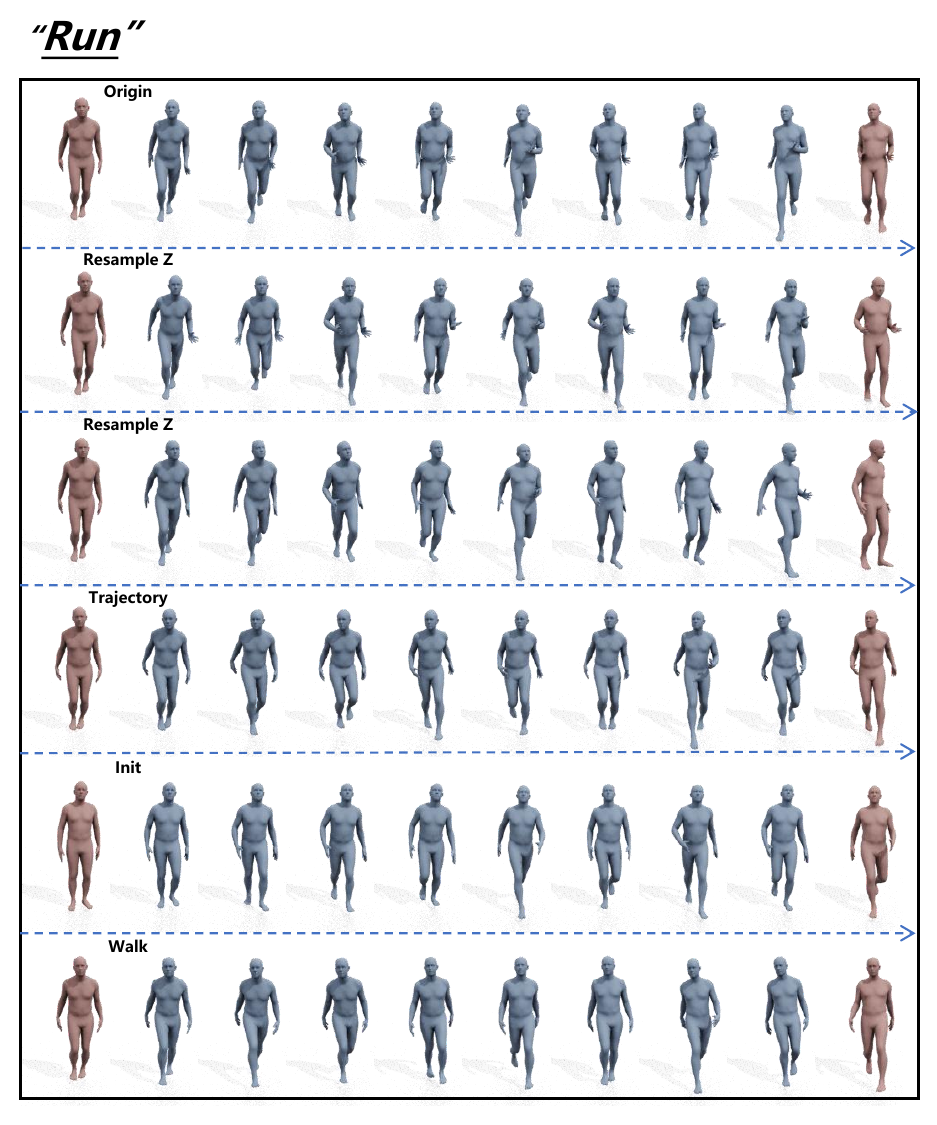}
    \vspace{-25pt}
    \caption{Effectiveness of different control signals. Note that, \emph{"Run"} in each row is downsampled to the same number of frames including the original start and end poses, and they are marked as the same color \emph{red} for better visualization.}
    \label{fig:singal_change2}
\end{figure}

In this section, we provide another example to show the effectiveness of control signals in Fig.~\ref{fig:singal_change2} and single-action visualization samples by sequences of dense frames in Fig.~\ref{fig:append:single_action_per_frame} \ as complements to the main text. Besides, more qualitative results of multi-action generation are included in supplementary videos for comprehensive evaluation.

\begin{figure*}[!ht]
    \centering
    \begin{overpic}[trim=3cm 6cm 1.5cm 0.5cm,clip,width=1\linewidth,grid=false]{figures/res2_singleAction_perframe.jpeg}
    \put(5,79){\footnotesize \textbf{"jump"}}
    \put(5,69.5){\footnotesize \textbf{"cross limbs"}}
    \put(5,59.5){\footnotesize \textbf{"stand up"}}
    \put(5,49.5){\footnotesize \textbf{"kick"}}
    
    \put(5,40){\footnotesize \textbf{"walk upward"}}
    \put(5,30){\footnotesize \textbf{"stretch"}}
    \put(5,20){\footnotesize \textbf{"move sth."}}
    \put(5,10){\footnotesize \textbf{"sit down"}}
    \end{overpic}\vspace{-20pt}
    \caption{Dense-frames visualization for eight synthesized actions using {\thename}. Each action is down-sampled to the same number of frames for better visualization.}
    \label{fig:append:single_action_per_frame}
\end{figure*}

    

\section{Application}
%

\subsection{System Workflow}
\begin{figure}[!h]
    \centering
    \vspace{-8pt}
    \includegraphics[width=1\linewidth]{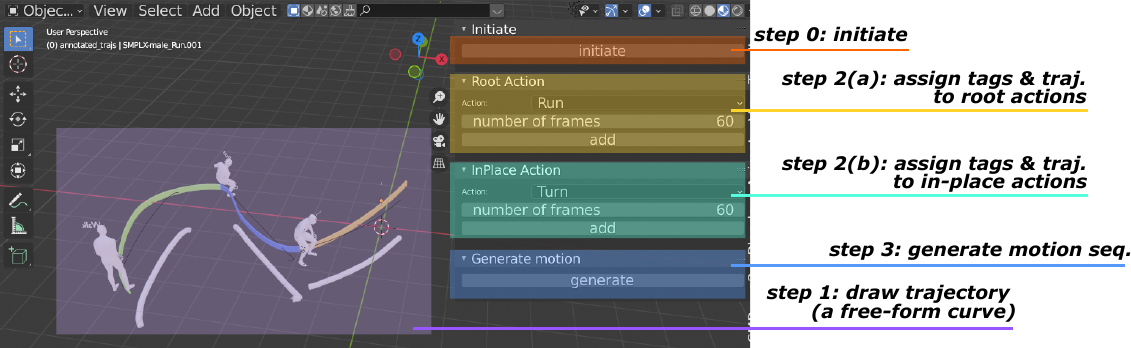}\vspace{-8pt}
    \caption{Screenshot of our trajectory annotation GUI.}\vspace{-8pt}
    \label{fig:system:ui}
\end{figure}

\begin{figure*}[!ht]
    \centering
    \includegraphics[width=0.8 \textwidth]{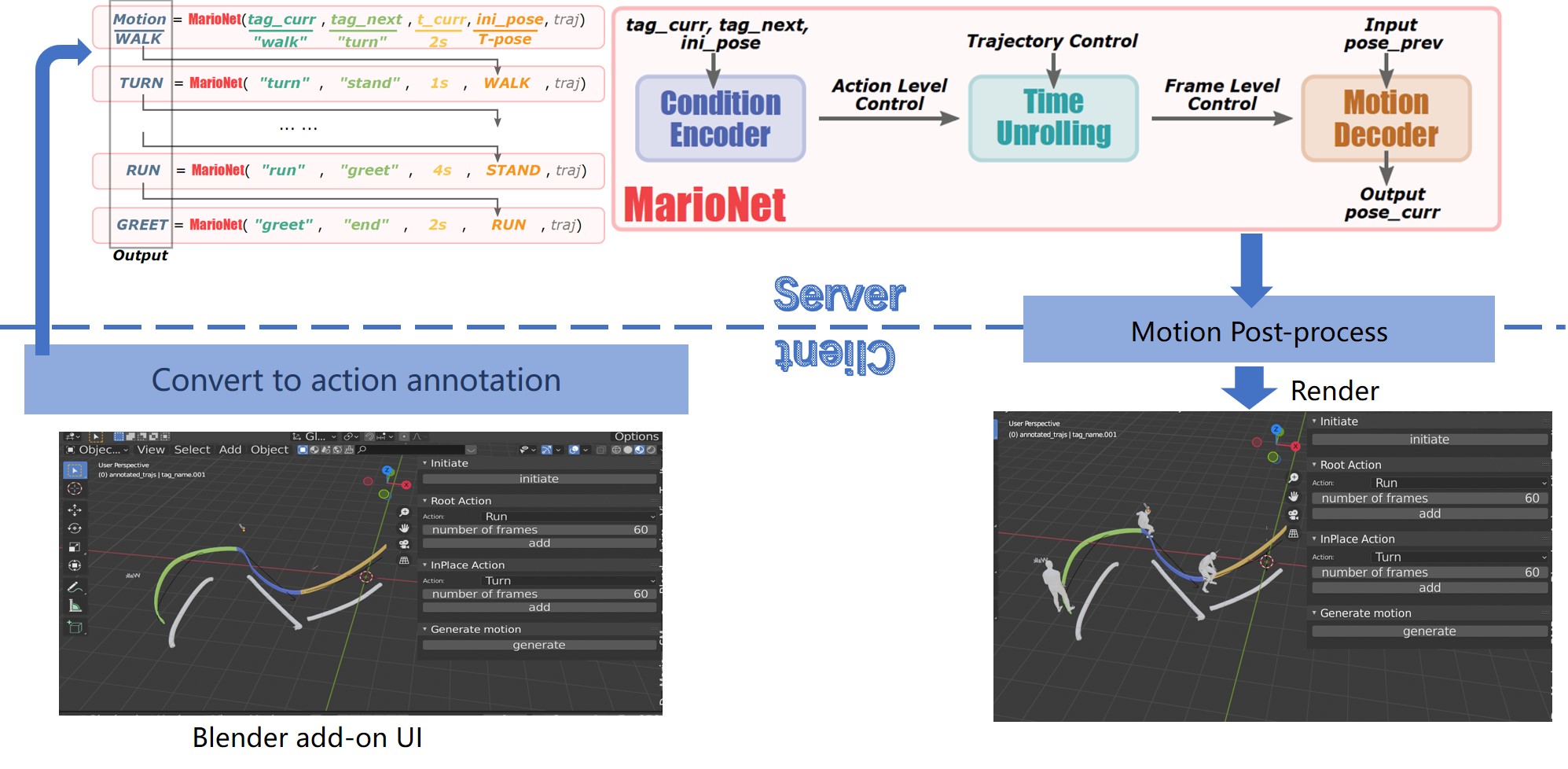}
    \caption{Application structure. Our interactive multi-action generation system is a client-server application. The user creates action annotations on the client side and sends them to the server while the server runs \systemname~ to synthesize motions and send them back to display on the screen.}
    \label{fig:system_structure}
\end{figure*}

Our interactive motion generation system is a client-server application (see Fig.~\ref{fig:system_structure}). It includes three main parts:  
(1) a user interface 
(see Fig.~\ref{fig:system:ui}) 
that allows users to a draw free-form curve to specify motion trajectory, select multiple action tags with the corresponding duration, and support displaying multiple generated motions;
(2) a pre-processing step to convert the hand-drawn trajectory into proper inputs of our network;
(3) a multi-action synthesis step utilizing {\systemname} 
to generate the whole motion and return data to the interface.

The users can follow the steps below to generate expected motions. And a user case of human motion generated in the 3D scene is demonstrated in Fig. 13 in the main text.

\begin{enumerate}
    \item Scene Preparation (Optional). Prepare or import a scene to display the expected motion.

    \item Trajectory Drawing. Click the "Initiate" button, then draw a free-form curve by clicking on the canvas, which will be taken as the motion trajectory.

    \item Per-action annotation. If necessary, one can insert points into the curve to divide it into segments for different actions by clicking. 
    Then select a curve segment (or a point) to assign tags and duration for a root action (or an in-place action) and click "add" to finish the current assignment. 
    Note that, a path (or a circle) will be displayed to show the added root (or in-place) action. One can repeat this step to finish multiple action assignments. 
     As for the in-place actions at the same circle, the action order in the synthesized motion is the same as the annotation order.
    \item  Generation. Click the "Generate" button to start the motion synthesis and check the results displayed on the canvas.
    \end{enumerate}
    
\subsection{Annotation Pre-processing}
\label{appendix:preprocess}
Firstly, the initial motion of the first action is either provided by the user or randomly selected from the training set conditioned on the same current tag. As for that in the following action, it is obtained by collecting the last 6 frames from the previously generated action.
The current tag, the following tag, and the duration for each action can be obtained from the user annotations directly. 
Note that, the 'following tag' is simply assigned as the corresponding 'current tag' for the last action by default.

The trajectory for an in-place action is set to zero value since no global translation of the root node is expected. 
For a root action, we fit a cubic Bézier curve
%
The positions of the samples are used as trajectory locations. 
Specifically, we have the \emph{trajectory location} for the $n$-th frame as follows:
\begin{equation}\label{eq:system:bezier}
\Scale[0.85]{ \mathbf{B(t)} = (1-t)^3 P_0 + 3(1-t)^2 t P_1 + 3(1-t)t^2 P_2 +t^3P_3 }
\end{equation}
where \(t = \frac{n}{T}\), $T$ is the total number of frames, and $(P_0, P_1, P_2, P_3)$ are the positions of the Bézier points during the fitting processing.
Finally, the body orientation of normalized BOP and the direction of processed trajectory for the first 6 frames would be checked whether their angle is out of the range of those in the training dataset. 
If an invalid angle appears, the body orientation would be automatically rotated to obey the rule.

\bibliographystyle{IEEEtran}
\bibliography{newReference}
%







